\documentclass[10pt,twocolumn,letterpaper]{article}

\usepackage[applications]{wacv}      

\usepackage{epsfig} 
\usepackage{mathptmx} 
\usepackage{times} 
\usepackage{amsmath} 
\usepackage{amssymb}  
\usepackage{algorithm2e}

\usepackage{balance}
\usepackage[utf8]{inputenc}
\usepackage{enumitem}
\usepackage[accsupp]{axessibility}
\usepackage{subcaption}
\usepackage{cuted}
\usepackage{multirow}
\usepackage{float}

\usepackage{tikz}

\usepackage{slashbox}

\usepackage{array}
\newcolumntype{?}{!{\vrule width 2pt}}

\newcommand{\uproman}[1]{\uppercase\expandafter{\romannumeral#1}}

\usepackage{soul}
\usepackage[textsize=tiny]{todonotes}

\usepackage{graphics} 
\usepackage{epsfig} 
\usepackage{mathptmx} 
\usepackage{times} 
\usepackage{amsmath} 
\usepackage{amssymb}  
\usepackage{xcolor}
\usepackage{algorithm2e}

\newcommand\blfootnote[1]{%
	\begingroup
	\renewcommand\thefootnote{}\footnote{#1}%
	\addtocounter{footnote}{-1}%
	\endgroup
}

\definecolor{wacvblue}{rgb}{0.21,0.49,0.74}
\usepackage[pagebackref,breaklinks,colorlinks,allcolors=wacvblue]{hyperref}
\usepackage[inkscapelatex=false]{svg}
\usepackage[absolute,overlay]{textpos}

\title{ZebraPose: Zebra Detection and Pose Estimation using \textit{only} Synthetic Data}

\author{Elia Bonetto$^{1,2}$\thanks{The authors thank the International Max Planck Research School for Intelligent Systems for supporting Elia Bonetto.} \and Aamir Ahmad$^{2,1}$}
\begin{document}

		\begin{textblock*}{\paperwidth}(0in,0.75in)
			\centering
			\parbox{0.8\textwidth}{\centering This paper has been accepted for publication in the \\ IEEE/CVF Winter Conference on Applications of Computer Vision 2026.\\
				Please cite as: Bonetto, E. and Ahmad, A. (2026). ZebraPose: Zebra Detection and Pose Estimation using \textit{only} Synthetic Data. \textit{IEEE/CVF Winter Conference on Applications of Computer Vision 2026 (WACV 2026)}.}
		\end{textblock*}
		
		\maketitle
		\begin{abstract}
Collecting and labeling large real-world wild animal datasets is impractical, costly, error-prone, and labor-intensive. For animal monitoring tasks, as detection, tracking, and pose estimation, out-of-distribution viewpoints (e.g. aerial) are also typically needed but rarely found in publicly available datasets. To solve this, existing approaches synthesize data with simplistic techniques that then necessitate strategies to bridge the synthetic-to-real gap. Therefore, real images, style constraints, complex animal models, or pre-trained networks are often leveraged. In contrast, we generate a fully synthetic dataset using a 3D photorealistic simulator and demonstrate that it can eliminate such needs for detecting and estimating 2D poses of wild zebras. Moreover, existing top-down 2D pose estimation approaches using synthetic data assume reliable detection models. However, these often fail in out-of-distribution scenarios, e.g. those that include wildlife or aerial imagery. Our method overcomes this by enabling the training of both tasks using the same synthetic dataset. Through extensive benchmarks, we show that models trained from scratch exclusively on our synthetic data generalize well to real images. We perform these using multiple real-world and synthetic datasets, pre-trained and randomly initialized backbones, and different image resolutions. Code, results, models, and data can be found at \href{https://zebrapose.is.tue.mpg.de/}{\url{https://zebrapose.is.tue.mpg.de/}}.
\end{abstract}    
		\section{Introduction}
\label{sec:intro}

\blfootnote{\hspace{-15pt}$^1$Max Planck Institute for Intelligent Systems. Max-Planck-Ring, 4, Tübingen, 72076, Germany. {\tt\small elia.bonetto@tue.mpg.de}\\$^2$University of Stuttgart, Faculty of Aerospace Engineering and Geodesy, Institute of Flight Mechanics and Control (iFR), Flight Robotics and Perception Group (FRPG). Pfaffenwaldring 27, 70569 Stuttgart, Germany. {\tt\small aamir.ahmad@ifr.uni-stuttgart.de}}

\begin{figure}[t]
    \centering
    \subcaptionbox{Elephant labeled as a zebra}{\includegraphics[width=0.48\columnwidth]{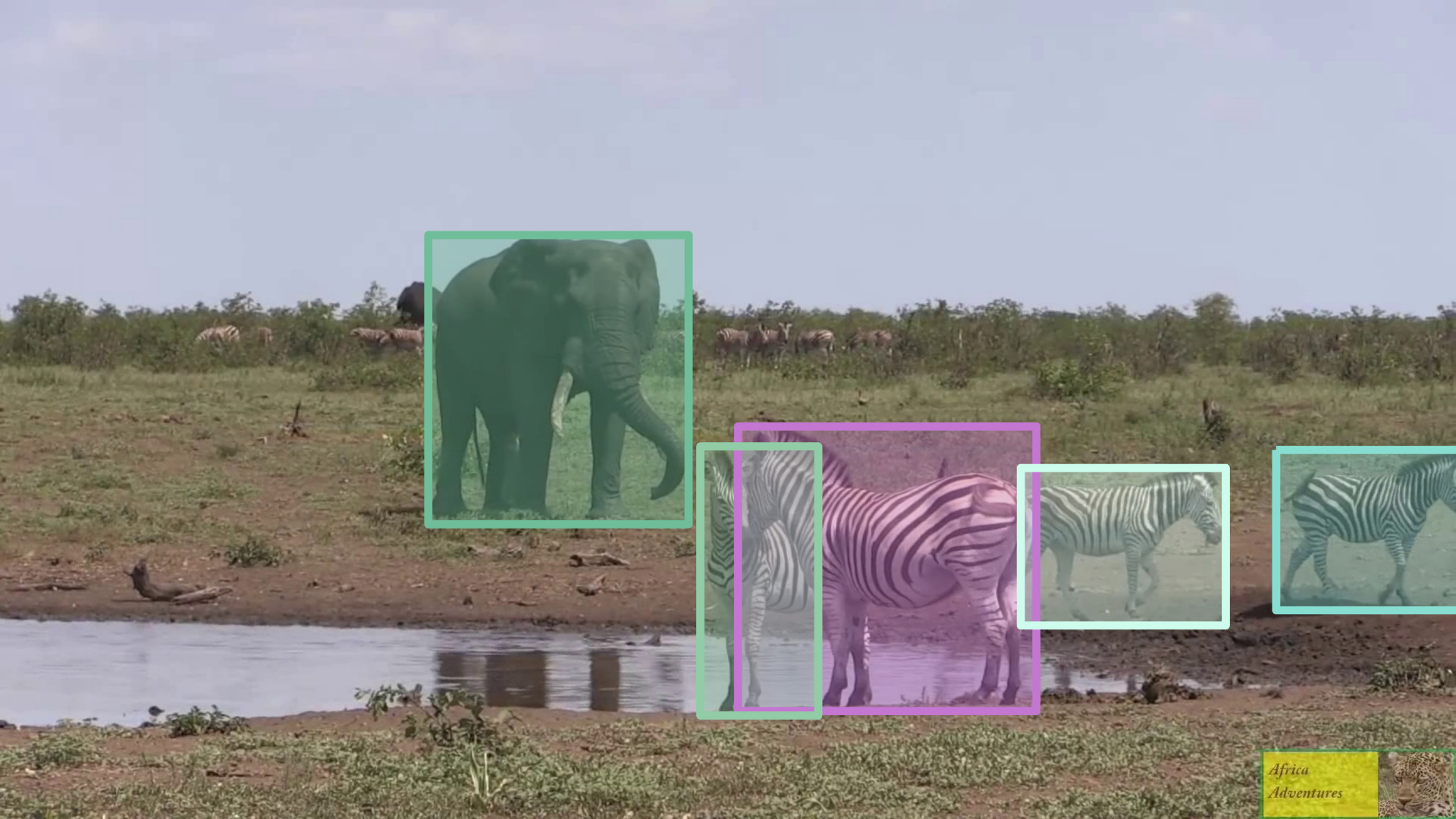}}\hfill
    \subcaptionbox{Two unlabeled individuals}{\includegraphics[width=0.48\columnwidth]{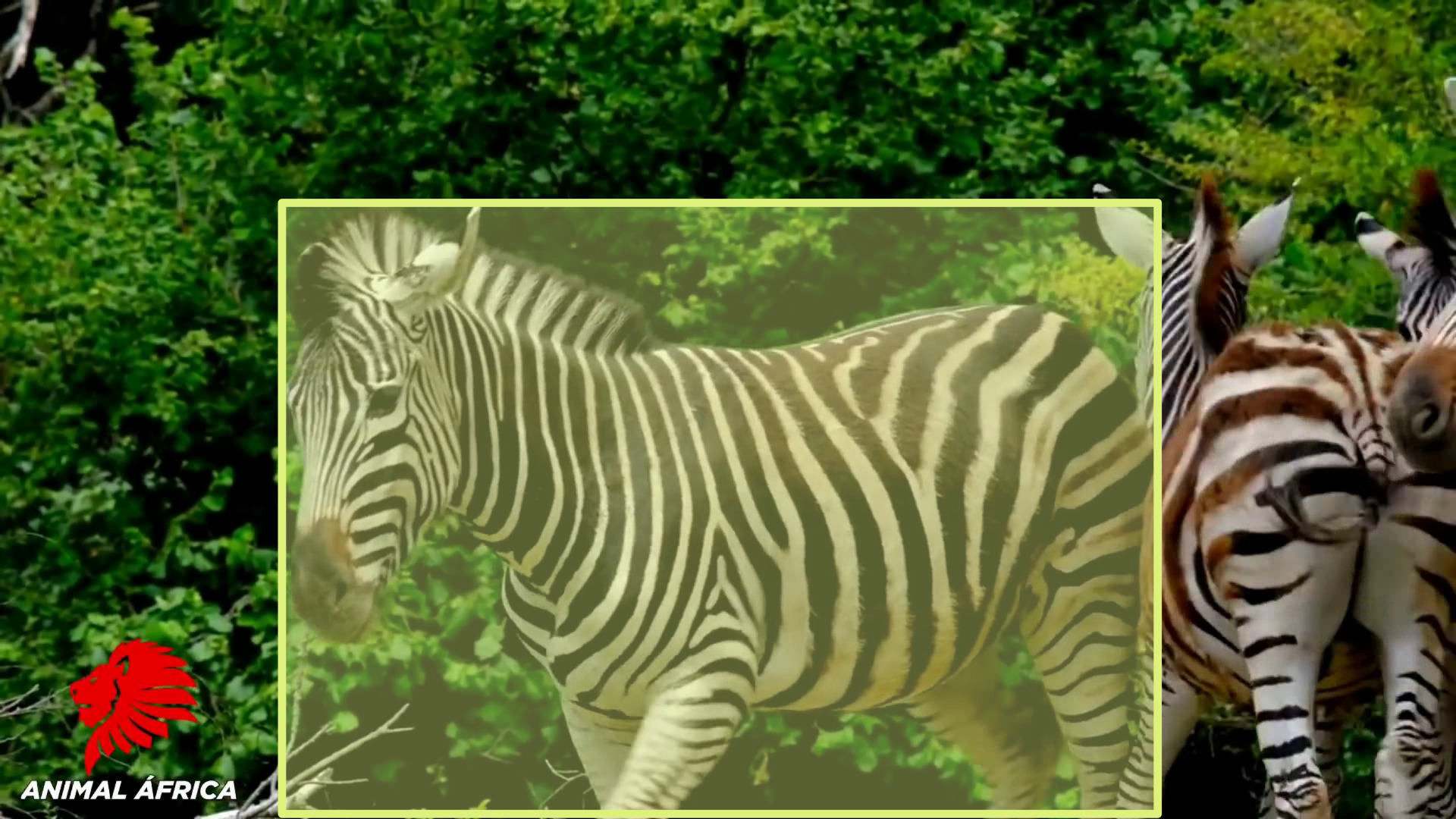}}\hfill
    
    \subcaptionbox{Bounding box is not tight}{\includegraphics[width=0.48\columnwidth]{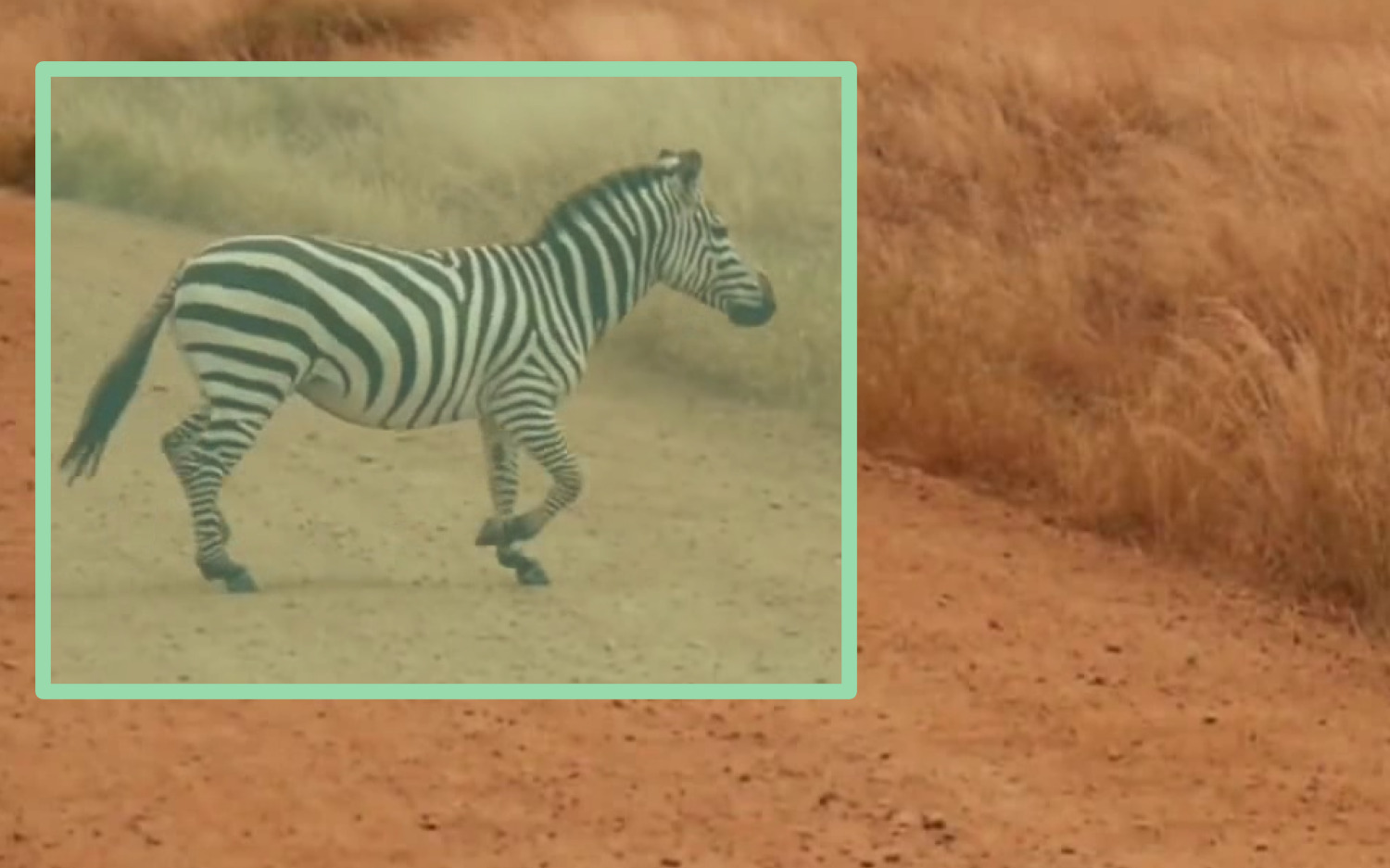}}\hfill
    \subcaptionbox{Bounding boxes are not tight}{\includegraphics[width=0.48\columnwidth]{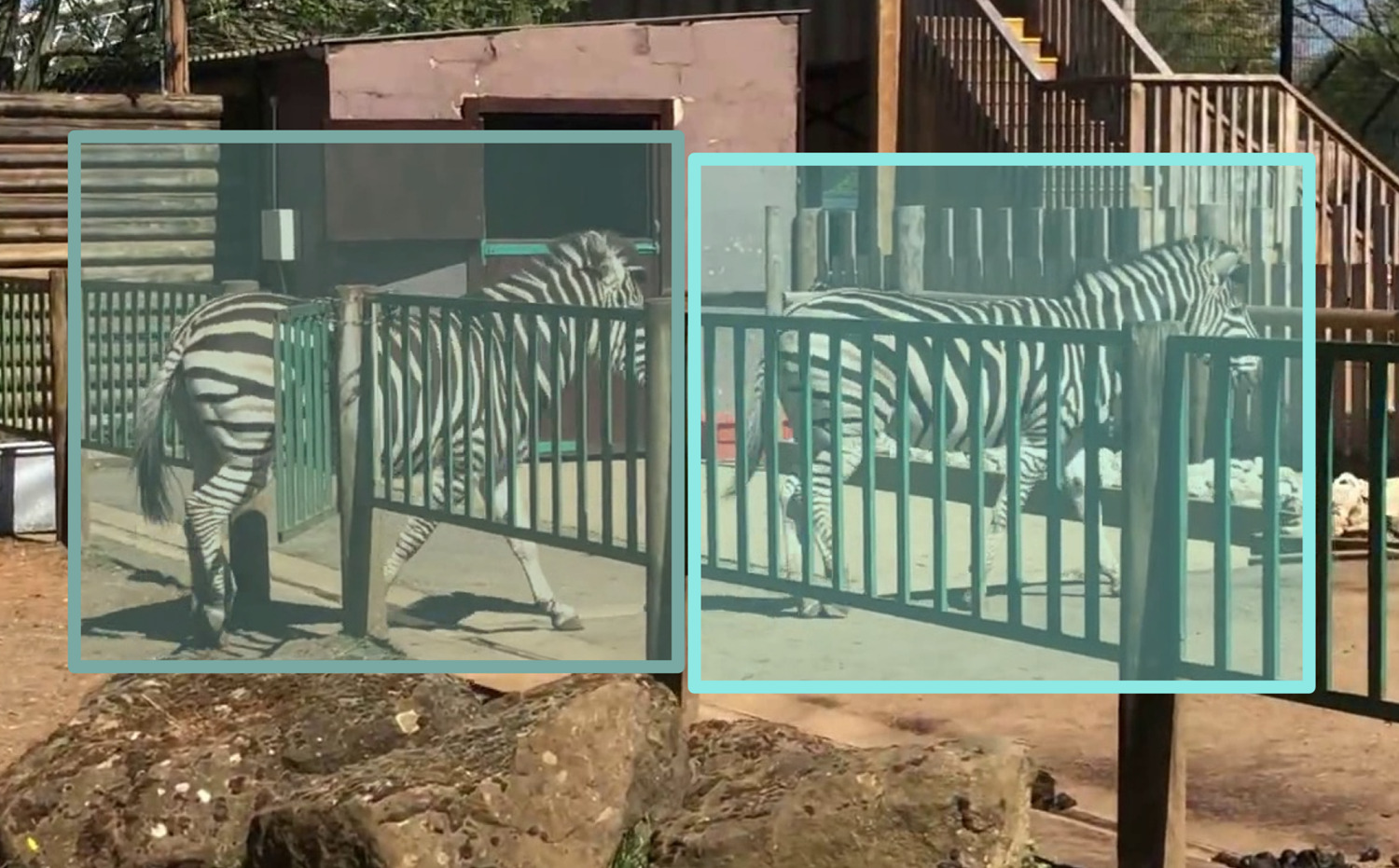}}\hfill
    
    \caption{Examples of annotation errors in the APT-36K dataset.}
    \label{fig:img_compare}
\end{figure}

Research in unconventional domains such as wildlife monitoring, animal pose estimation, and aerial image analysis is often hindered by the lack of labeled data~\cite{pasyn,spacnet,surveyKPs,bonetto-syn-zebras}. Applications in human pose estimation can leverage large-scale datasets obtained in controlled environments and motion capture systems~\cite{amass}. Unlike that, obtaining diverse and accurately labeled animal data remains a significant challenge. Manual annotation is costly, time-consuming, and prone to errors (\cref{fig:img_compare}). While semi-automatic systems such as VICON halls have been widely used in the past (e.g.~\cite{higami2024ratbodyformer,marshall2021continuous,descriptorposes,nagy2023smart}), they are often infeasible for wildlife or outside specialized labs. This is particularly true for endangered or large mammals (e.g. zebras, pandas), where fur, body size, and welfare and ethical concerns can make correct data collection or their confinement unfeasible. For example, wild creatures cannot be easily restrained without inducing stress, altering their behavior, or removing them from their natural habitats~\cite{Trondrud2022,Breed2019}. In this context, aerial-based detection and pose estimation of wild animals provide valuable insights by enabling non-intrusive monitoring of their health, motion patterns, and interactions with their natural environment~\cite{Zuffi2019ThreeDSL}.
\begin{figure}[!t]
    \centering
    \includegraphics[width=\columnwidth]{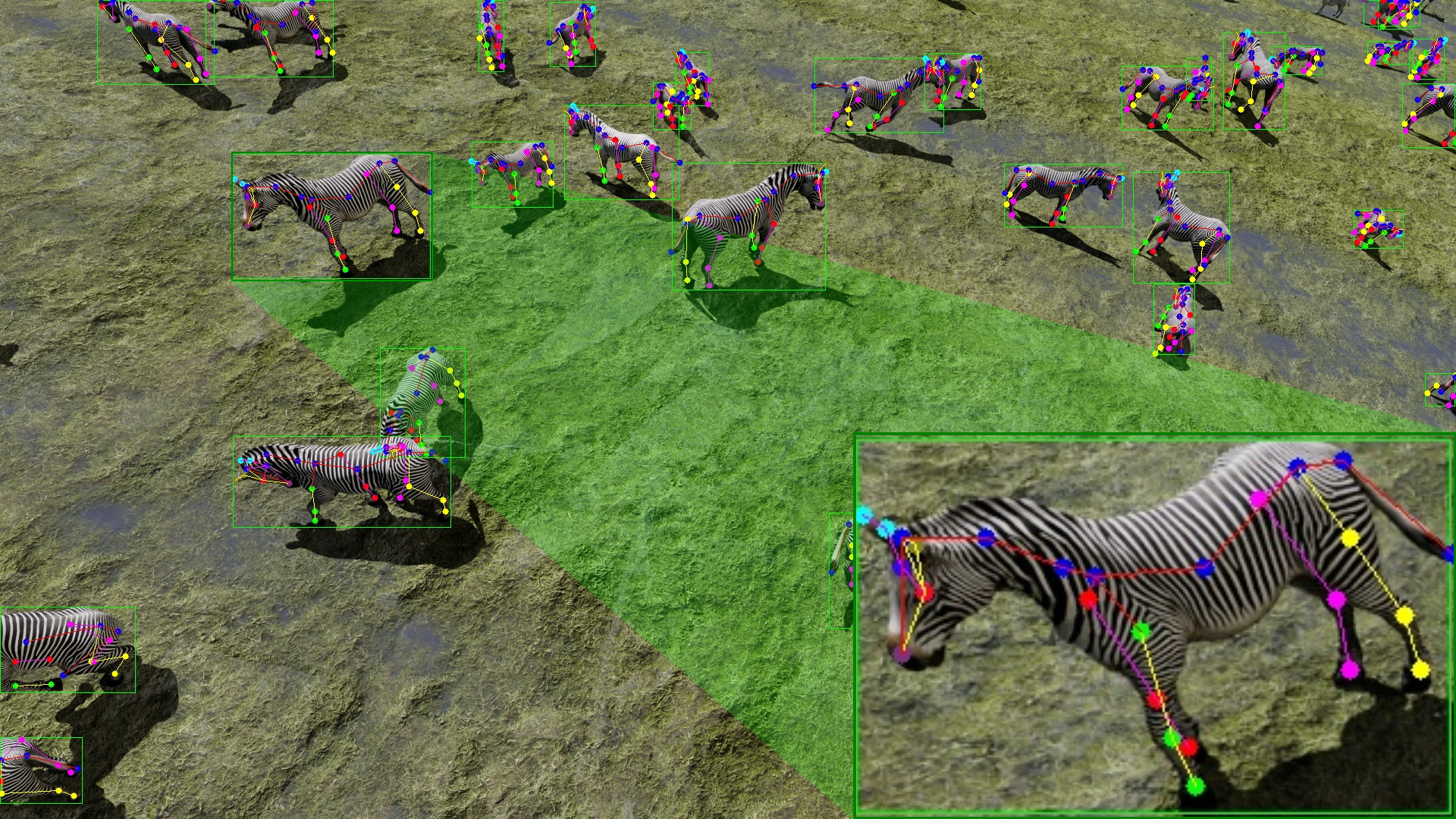}
    \caption{A sample of our synthetic data. Zoomed inset: an individual with all the 27 keypoints labeled.}
    \label{fig:cover}
\end{figure}
While detection is essential for localizing animals, pose estimation facilitates activity recognition and serves as a prior for shape estimation and health assessment~\cite{bray1999body}. However, the availability of animal-focused datasets collected in the wild remains scarce. Additionally, many existing ones lack aerial perspectives and out-of-distribution viewpoints~\cite{bonetto-syn-zebras,beery2020synthetic}. This domain gap remains a critical bottleneck for conservation-focused deep learning (DL) applications, limiting their real-world in-the-wild applicability. The challenge lies in acquiring diverse, high-quality datasets that enable DL models to generalize effectively across different environments and observation points. This is particularly pressing for species like Grévy's zebra, i.e. the focus of this work,  an endangered herbivore whose population dynamics can influence entire ecosystems~\cite{Smith2022}.

A promising approach to overcoming these limitations is synthetic data~\cite{uda,smal}. Synthetic datasets enable precise annotations, controlled environmental conditions, and the generation of large-scale, diverse samples. However, while both detection and pose estimation can benefit from synthetic data, their requirements differ. Detection performance depends on background realism, camera-subject distance, and viewpoint variability, whereas pose estimation relies on capturing intra-body relationships and depth cues~\cite{surveyKPs,peoplesanspeople}. Generating effective synthetic datasets thus requires both (i) an automated procedure to cover the target data distribution and (ii) strategies to bridge the synthetic-to-real (syn-to-real) gap. Many existing approaches address only one of these aspects, often simplifying backgrounds or relying on domain adaptation techniques such as style transfer or real-image supervision~\cite{bonetto-syn-zebras,beery2020synthetic,uda}. As a result, achieving robust real-world generalization using purely synthetic data remains an open challenge~\cite{surveyKPs}.

In this work, we propose a unified approach for \textit{both} detection and 2D pose estimation of zebras using \textit{only} synthetic data, i.e. a full top-down method. Unlike prior pose estimation works that often assume a pre-trained detector, we show that detection itself can be a bottleneck, e.g. in aerial settings (\cref{sec:detection_results}). Our approach leverages a 3D photorealistic simulator to generate a large-scale synthetic dataset and uses it to train \textit{both} the detection and pose estimation models. First, we extend the work of~\cite{bonetto-syn-zebras} by significantly increasing dataset variability through systematic image cropping, scaling, and augmentation, thereby improving generalization to non-aerial images (e.g. the ones in \cref{fig:img_compare}). We use this data to train YOLOv5s models for zebra detection and evaluate them on (i) existing real-world datasets and (ii) our newly introduced high-resolution dataset of 104K precisely labeled aerial images. Second, we automatically annotate all synthetic animals with 27 ground-truth keypoints extracted from their 3D meshes and train a ViTPose+~\cite{vitposeplus} model using both pre-trained and randomly initialized backbones. Extensive benchmarking confirms the generalization of our models to real-world zebra imagery. We further demonstrate that minimal real data enables effective adaptation to horse pose estimation. In summary, our main contributions are:
\begin{itemize}[leftmargin=*]
    \item A \emph{generalized} detection and 2D pose estimation pipeline for zebras trained exclusively on synthetic data, validated through extensive benchmarking on real-world datasets.
    \item An in-depth study on the syn-to-real gap for both detection and pose estimation, analyzing how dataset variability and augmentation improve generalization across \textit{seven} different real-world datasets.
    \item A large dataset of zebras observed by UAVs and precisely labeled with bounding boxes.
\end{itemize}

		\section{Related Works}
\label{sec:soa}

Due to the diversity observed within the animal kingdom—even among closely related species such as \textit{Canidae} and the challenges in acquiring (pseudo)ground-truth data, it is difficult to construct a universal animal model analogous to what SMPL is for humans~\cite{smpl}. As a consequence, synthetic data generation for animals typically involves stitching onto randomized backgrounds CAD or SMAL-based models~\cite{smal,smalr} that are animated via manual manipulation, pre-fitting to real images, or posing VAEs~\cite{lsa,uda,cda,pasyn,spacnet}. While this approach enables rapid data generation, it often introduces scale inconsistencies, poor blending with the scene, and unrealistic lighting. Indeed, existing synthetic datasets are primarily designed for 2D pose estimation (i.e. keypoints estimation), focusing on joint orientation and positioning rather than overall visual fidelity~\cite{surveyKPs}. This limitation, coupled with the absence of robust priors, leads many studies to face real-world generalization issues. To mitigate the syn-to-real gap, researchers frequently rely on domain adaptation or semi-supervised learning techniques—leveraging large quantities of unlabeled real images or enforcing consistency, style, and other constraints during data generation~\cite{lsa,spacnet,uda,li2023scarcenet}. However, despite some performance gains from incorporating unlabeled data, these methods often suffer from overfitting and noise in the refinement of synthetic data predictions~\cite{pasyn,surveyKPs,li2023scarcenet}, and they are typically evaluated on only a limited set of datasets. Moreover, while recent approaches such as~\cite{clamp,groundingDino} have integrated large language models (LLMs) with synthetic data to address common issues like left/right flipping, their performance still lags behind state-of-the-art models such as ViTPose+ \cite{vitposeplus}. Recent work on human pose estimation \cite{bedlam} further highlights the importance of photorealism in bridging the syn-to-real gap.

A critical limitation of current keypoints estimation methods using synthetic data is the underlying assumption that detection is already solved~\cite{surveyKPs,pasyn}. Most synthetic datasets consist of tightly cropped images around the animal~\cite{lsa,uda}, which speeds up rendering and yields crisp images ideal for pose estimation. However, this cropping renders the data unsuitable for training detection models since the animal always occupies the full image and occlusions are rare. Therefore, most studies do not evaluate the joint task of detection and pose estimation using synthetic data, assuming off-the-shelf high-quality detections. Notably, only~\cite{bonetto-syn-zebras} has explored detecting animals directly from synthetic images. In their approach, they generated visually appealing data using easily obtainable models within a photorealistic simulator~\cite{GRADE}. However, their randomized viewpoint strategy predominantly produces aerial perspectives, limiting its utility for standard, ground-level detection tasks. Moreover, they do not evaluate the impact of image resolution in either training or evaluations, although that might significantly impact performance. In this work, we build upon the dataset from~\cite{bonetto-syn-zebras} and extend it to address both detections in general images and pose estimation.

A parallel line of work explores AIGC-based dataset generation (e.g., AniMer~\cite{lyu2025animer}, GenZoo~\cite{niewiadomski2024generative}), which improves visual fidelity and scalability. While promising in terms of realism and variability, these approaches often lack deterministic control over key parameters such as camera settings, pose, and environment. In contrast, simulators like IsaacSim enable physics-based rendering and seamless integration with robotics platforms, which is essential for extending methods across domains and testing new approaches. Notably, recent work on inverse graphics~\cite{kulitsre,kulits2025reconstructing} suggests that combining generative models with simulation and LLM may bridge this gap, enabling the creation of simulation-ready worlds and pointing to a promising future direction.
		\section{Approach}
\label{sec:approach}
Our goal is to obtain a \textit{full} top-down system for the detection and 2D pose estimation of zebras in real images using \textit{only} synthetic data for training. We use the popular YOLOv5~\cite{yolov5} for detecting the animals, and ViTPose+~\cite{vitposeplus} for estimating the 2D keypoints. To generate synthetic data, we build upon~\cite{bonetto-syn-zebras}, which uses the GRADE framework~\cite{GRADE} and introduces into it an animated zebra model\footnote{\url{https://skfb.ly/opCUB}} with ten different environments from the Unreal Engine marketplace. As this is a freely available asset, it has only one texture and shape state, and a limited pose variability. Regardless, given enough generated samples, we can train both effective detection and pose estimation models, as shown later. Using these, 250 randomly scaled and posed 3D zebra models are placed in the simulated environment, ensuring no collisions. UAVs are positioned based on the zebras’ average 3D location, and scenes are rendered at 1080p. Their resulting `SC' dataset contains 18K annotated frames with RGB images, bounding boxes, instance masks, depth, and vertex locations. However, detection models trained only on SC data fail in common images, e.g. on the ones from APT-36K~\cite{bonetto-syn-zebras}. Since they rely on a photorealistic simulator and realistic assets and achieve good performance in the detection from aerial views, we can assume a good general quality of the data in terms of photorealism. We thus argue that the failure to generalize to common images is related to the actual distribution of the viewpoints and the relation between the size of the individuals and images during testing. Indeed, since cameras' locations are uniformly randomized, the number of times they are near the animals is much lower than when they are far from them. Moreover, the training and validation input dimension of YOLOv5 can impact the performance of the network itself, as we show in~\cref{sec:detection_results}, since scaling influences both the size of the objects in the scene and the visual quality. For example, a 90px wide box in a 1920$\times$1080 image when scaled to a $640\times480$ image becomes three times smaller. Notably, the zebra animal model can be exchanged with other assets (e.g.~\cite{3danimalmodels,gimstudio}) without modifying the pipeline, allowing the extension of this work to different species.

\subsection{Detection}
\label{sec:detection_aug}
\begin{figure}[!ht]
    \centering
    {\includegraphics[width=.49\columnwidth]{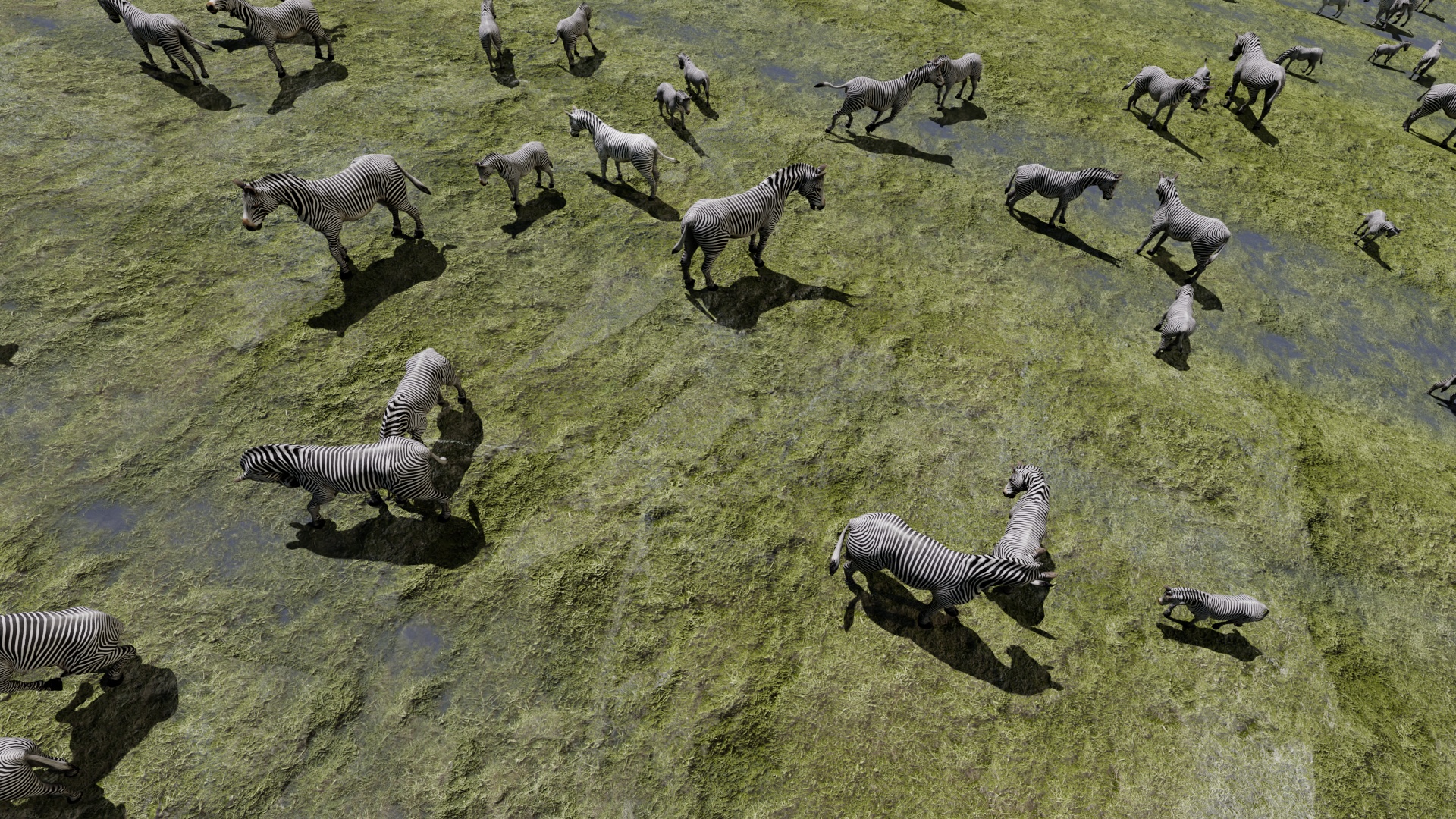}}\hfill
{\includegraphics[width=.49\columnwidth]{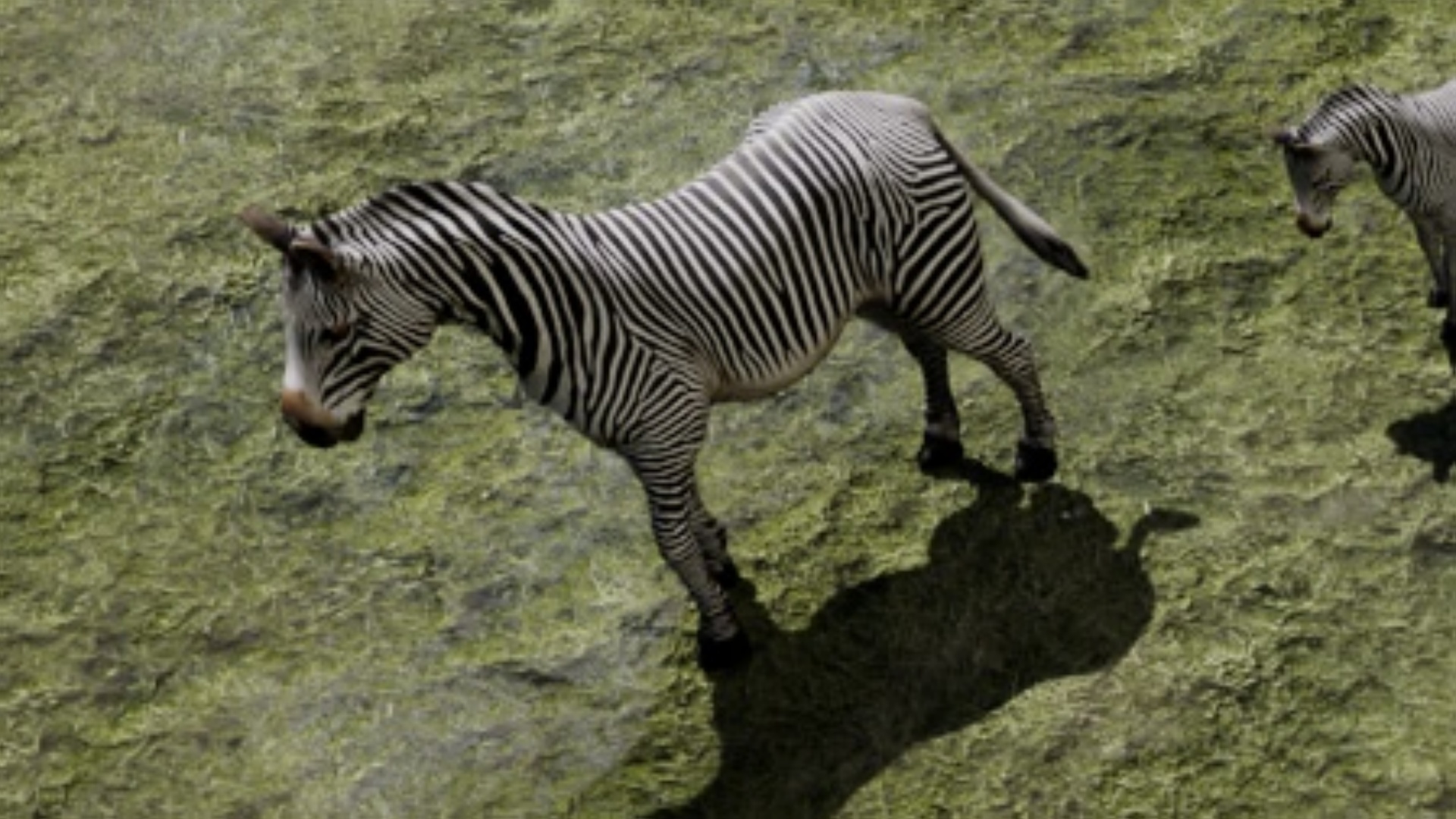}}\hfill
    \caption{An example of before [left] and after [right] the cropping and scaling procedure (\Cref{sec:detection_aug}).}
    \label{fig:upscale}
\end{figure}

\begin{figure}
    \centering
    \includegraphics[width=0.46\columnwidth]{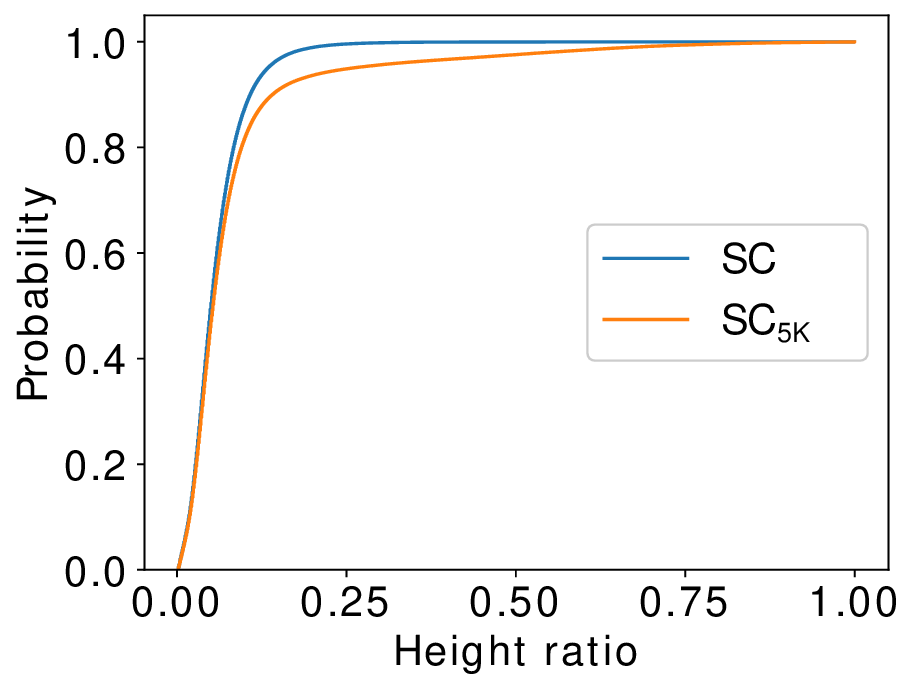}
    \includegraphics[width=0.46\columnwidth]{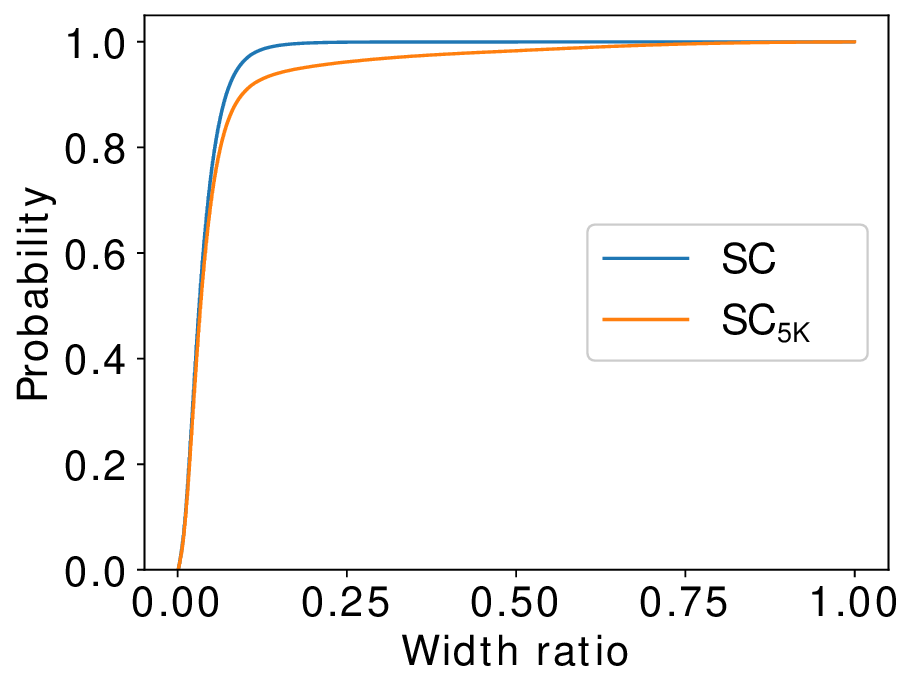}
    \caption{Cumulative Distributions of height and width ratio w.r.t. image size for SC and SC$_{\text{5K}}$ datasets.}
    \label{fig:cdfs}
\end{figure}

Following the intuitions above, instead of generating new synthetic data by creating a new pipeline, we introduce a new pre-processing step to augment the `SC' dataset to obtain wider coverage of the desired distribution. Since the original dataset has been rendered from random viewpoints, the individuals in the images tend to be small. Therefore, to create images with bigger (virtually `closer') animals, we employ a new targeted crop and scale procedure. We first select all the animals whose bounding box area is greater than a given threshold for each image in the training and validation sets. Each box is randomized using a random offset between 0 and 150 pixels on each side, creating a rectangular area around the animal. We then crop this box and rescale it to the original image size (i.e. $1920\times1080$). The annotations are finally generated from the upscaled ground-truth segmentation masks of the crop itself. Note that the upscaling operation degrades the quality of the final image, as it acts as a digital zoom, and we do not re-render the scene. We set the threshold on the original bounding box area to 5000 pixels to filter excessively small animals. An example of a zoomed-in individual is given in~\cref{fig:upscale}, while the variation of the cumulative density function of the boxes' dimensions is shown in~\cref{fig:cdfs}. The resulting data, identified as SC$_{\text{5K}}$, consists of 23K training and 5.8K validation frames, with an increment of 36K instances.

\subsection{2D Pose Estimation}
\label{sec:keypoints_gen}
The `SC' data does not contain keypoints information out of the box. To obtain that, we exploited the information at our disposal, i.e. the known 3D location of the camera and the 3D vertex locations for each individual. We create 27 different groups of vertices, corresponding to 27 different keypoints. The resulting keypoints, depicted in~\cref{fig:cover}, are: four hoofs, four knees, four thighs, the tail start and end locations, eyes, ear tips and bases, start and end of the neck, nose, skull, body middle, and back end and front. From each set of vertices, we compute the 3D average location and project it to the image. Using the COCO convention, we mark as \textit{visible} the keypoints within the \textit{instance} mask of the animal, and as \textit{occluded} the remaining ones. We extract keypoints directly from the stored per-frame 3D mesh representations, enabling pixel-perfect, consistent annotations \textit{without} re-rendering. The process is fully customizable as keypoints can be redefined by updating the annotation logic, e.g. by using skeletal joints. We label only the individuals whose maximum bounding box dimension, either width or height, exceeds 30 pixels. As we use a top-down approach, each annotation is separately cropped and scaled to the input size of the selected model and only then used for training and validation, with the loss computed only on the labeled keypoints. This automatically diminishes the impact of the relative distance between the camera and the animal, while influencing the sharpness of the final crop. Finally, we map the 17 canonical keypoints of datasets such as APT-36K to our 27 to train models using mixed data. Note that the whole set of animations of the synthetic zebra model consists of only \textbf{888} different poses. Out of those, \textbf{440} are \textit{idling} (practically static) ones; a rather limited number of them if compared to other datasets~\cite{spacnet,lsa}.
		\section{Experiments and Evaluations}
\label{sec:exp}

\subsection{Datasets}
In the next sections, we use the short names of the datasets introduced here (in bold). When mixed datasets are used during training procedures, we will concatenate the names using the `+' symbol. A recap of the datasets used is reported in~\cref{tab:datasets} in the supplementary material.

\noindent\textbf{Synthetic data:}
\textbf{SC} is the synthetic data from~\cite{bonetto-syn-zebras}, which consists of 18K frames divided with an 80/20 split between training and validation sets. SC has been used to generate both the keypoints (\cref{sec:keypoints_gen}) annotations and the augmented dataset (\cref{sec:detection_aug}), i.e. \textbf{$\text{SC}_{\text{5K}}$}. $\text{SC}_{\text{5K}}$ consists of 29K images divided again with an 80/20 ratio. We also use the zebra images from the \textbf{SpacNet}~\cite{spacnet} data, consisting of 3000 generated frames with style transfer applied to them. We use this to evaluate the quality of the synthetic data generated on the 2D pose estimation task, comparing a vanilla generation with a more complex method (fully described in~\cite{spacnet}).

\begin{table*}[!ht]
    \centering
    \resizebox{.85\textwidth}{!}{%
    \begin{tabular}{l|cc|cc|cc|cc|cc|cc}
    \toprule[1.5pt]
        \multicolumn{1}{r|}{Val. Set $\rightarrow$} & \multicolumn{2}{c|}{$\text{A36}_\text{OZ}$}  & \multicolumn{2}{c|}{$\text{A10}_\text{OZ}$}  &  \multicolumn{2}{c|}{RP}  & \multicolumn{2}{c|}{R123} & \multicolumn{2}{c|}{Average} & \multicolumn{2}{c}{W. Avg.}\\ \cline{2-13}
         
         Train Set $\downarrow$& mAP50 & mAP & mAP50 & mAP & mAP50 & mAP & mAP50 & mAP & mAP50 & mAP & mAP50 & mAP \\ \hline
SC & 0.150 & 0.053 & 0.076 & 0.021 & 0.331 & 0.228 & 0.911 & 0.611 & 0.367 & 0.228 & 0.899 & 0.603 \\
{$\text{SC}_{\text{5K}}$} & 0.512 & 0.197 & 0.268 & 0.094 & 0.340 & 0.236 & 0.935 & 0.623 & 0.514 & 0.288 & 0.928 & 0.617 \\
SC+RP & 0.207 & 0.102 & 0.084 & 0.030 & 0.903 & 0.638 & 0.980 & 0.683 & 0.544 & 0.363 & 0.969 & 0.675 \\
{$\text{SC}_{\text{5K}}$}+RP & 0.580 & 0.243 & 0.309 & 0.106 & 0.910 & 0.639 & 0.975 & 0.680 & 0.693 & 0.417 & 0.969 & 0.674 \\
SC+$\text{CZ}_{1920}$ & 0.709 & 0.386 & 0.466 & 0.234 & 0.350 & 0.253 & 0.922 & 0.625 & 0.612 & 0.374 & 0.918 & 0.621 \\
{$\text{SC}_{\text{5K}}$}+$\text{CZ}_{1920}$ & 0.740 & 0.447 & 0.517 & 0.277 & 0.370 & 0.268 & 0.940 & 0.638 & 0.642 & 0.407 & 0.936 & 0.634 \\ \hline
SC+$\text{CZ}_{1920}$+RP & 0.705 & 0.383 & 0.460 & 0.228 & 0.921 & 0.639 & 0.980 & \textbf{0.689} & 0.766 & 0.485 & 0.975 & \textbf{0.685} \\ 
{$\text{SC}_{\text{5K}}$}+$\text{CZ}_{1920}$+RP & 0.752 & 0.434 & 0.442 & 0.237 & \textbf{0.926} & \textbf{0.665} & \textbf{0.982} & 0.688 & \textbf{0.775} & \textbf{0.506} & \textbf{0.978} & 0.684 \\ \hline
$\text{CZ}_{1920}$ & \textbf{0.839} & \textbf{0.514} & \textbf{0.740} & \textbf{0.432} & 0.244 & 0.149 & 0.644 & 0.343 & 0.617 & 0.360 & 0.646 & 0.345 \\ 
$\text{CZ}_{640}$ & 0.496 & 0.269 & 0.220 & 0.116 & 0.441 & 0.311 & 0.856 & 0.561 & 0.503 & 0.314 & 0.850 & 0.556 \\ 
 \bottomrule[1.5pt]
    \end{tabular}
    }
    \caption{YOLOv5s models validated using images scaled to $1920\times1920$ pixels. We report mAP50 and mAP for each validation set (columns), compute the average and the weighted average (on the number of images), and bold the best metrics.}
    \label{tab:yolo-1920}
\end{table*}

\begin{table*}[!ht]
    \centering
    \resizebox{0.85\textwidth}{!}{%
    \begin{tabular}{l|cc|cc|cc|cc|cc|cc}
    \toprule[1.5pt]
                \multicolumn{1}{r|}{Val. Set $\rightarrow$} & \multicolumn{2}{c|}{$\text{A36}_\text{OZ}$}  & \multicolumn{2}{c|}{$\text{A10}_\text{OZ}$}  &  \multicolumn{2}{c|}{RP}  & \multicolumn{2}{c|}{R123} & \multicolumn{2}{c|}{Average} & \multicolumn{2}{c}{W. Avg.}\\ \cline{2-13}
                
     \multicolumn{1}{l|}{Train Set $\downarrow$}& mAP50 & mAP & mAP50 & mAP & mAP50 & mAP & mAP50 & mAP & mAP50 & mAP & mAP50 & mAP \\ \hline
SC & 0.613 & 0.343 & 0.455 & 0.245 & 0.158 & 0.116 & 0.581 & 0.382 & 0.452 & 0.271 & 0.580 & 0.380 \\
{$\text{SC}_{\text{5K}}$} & 0.676 & 0.361 & 0.583 & 0.318 & 0.164 & 0.119 & 0.605 & 0.408 & 0.507 & 0.302 & 0.605 & 0.407 \\
SC+RP & 0.690 & 0.414 & 0.552 & 0.315 & \textbf{0.471} & \textbf{0.256} & 0.734 & 0.502 & 0.612 & 0.372 & 0.732 & 0.500 \\
{$\text{SC}_{\text{5K}}$}+RP & 0.753 & 0.434 & 0.613 & 0.352 & 0.464 & 0.254 & 0.769 & 0.520 & 0.650 & 0.390 & 0.768 & 0.518 \\
SC+$\text{CZ}_{1920}$ & 0.827 & 0.495 & 0.889 & 0.601 & 0.163 & 0.119 & 0.565 & 0.375 & 0.611 & 0.398 & 0.568 & 0.377 \\
{$\text{SC}_{\text{5K}}$}+$\text{CZ}_{1920}$ & 0.846 & 0.521 & 0.917 & 0.632 & 0.171 & 0.124 & 0.646 & 0.436 & 0.645 & 0.428 & 0.648 & 0.437 \\\hline
SC+$\text{CZ}_{1920}$+RP & 0.861 & 0.534 & 0.894 & 0.603 & 0.447 & 0.244 & 0.731 & 0.500 & 0.733 & 0.470 & 0.732 & 0.500 \\ 
{$\text{SC}_{\text{5K}}$}+$\text{CZ}_{1920}$+RP & 0.872 & 0.545 & \textbf{0.911} & 0.631 & 0.459 & 0.247 & \textbf{0.770} & \textbf{0.524} & \textbf{0.753} & \textbf{0.487} & \textbf{0.771} & \textbf{0.524} \\\hline
$\text{CZ}_{1920}$ & 0.868 & 0.524 & 0.896 & 0.621 & 0.081 & 0.044 & 0.146 & 0.072 & 0.498 & 0.315 & 0.155 & 0.078 \\ 
$\text{CZ}_{640}$ & \textbf{0.888} & \textbf{0.575} & 0.903 & \textbf{0.646} & 0.208 & 0.135 & 0.411 & 0.249 & 0.602 & 0.401 & 0.417 & 0.253 \\
 \bottomrule[1.5pt]

    \end{tabular}
    }
    \caption{YOLOv5s models validated using images scaled to $640\times640$ pixels. We report mAP50 and mAP for each validation set (columns), compute the average and the weighted average (on the number of images), and bold the best metrics.}
    \label{tab:yolo-640}
\end{table*}

\noindent\textbf{Real-world data --- Common:} Contrary to previous works, we use multiple real-world datasets to evaluate our data and thoroughly test our method's generalization and flexibility. This is necessary as some datasets, such as \textbf{Zebra-300}~\cite{pasyn}, are simpler than others (\cref{sec:pose_results}). We adopt the \textbf{A}P-\textbf{10}K~\cite{ap-10k} dataset, named \textbf{A10} in this work. The full dataset consists of 10015 images of various species already divided into training, validation, and test sets. Like other prior works~\cite{surveyKPs, uda, lsa, spacnet}, we adopt the first split among the ones available. We subdivide A10 into \textbf{$\text{A10}_\text{OZ}$}, i.e. the subset containing \textbf{O}nly \textbf{Z}ebras, and \textbf{$\text{A10}_{99}$}, i.e. a random subset of \textbf{99} zebras (described in~\cite{pasyn}). We also use the \textbf{A}PT-\textbf{36}K~\cite{apt-36k} dataset, named as \textbf{A36}, consisting of 2400 videos, summing up to 35K frames, of different animals. Since an official split has not been released, we divide the dataset using an 80/20 ratio, keeping videos in either of the sets to avoid overlap. Again, we filter from A36 \textbf{O}nly the \textbf{Z}ebras labels, obtaining \textbf{$\text{A36}_\text{OZ}$}. Other zebras-only datasets used in this work are the \textbf{Zebra-300} and \textbf{Zebra-Zoo}~\cite{pasyn}, which have been used only for validation purposes. Zebra-300 contains 40 images from the AP-10K test set, 160 images from the AP-10K unlabeled set, and 100 from the Grévy's zebra~\cite{dpk-zebrads} dataset according to~\cite{pasyn}. To evaluate the generalization capabilities, we use the \textit{horse} subset of the \textbf{T}ig\textbf{D}og~\cite{tigdog} dataset, which we called \textbf{TDH}. We use the preprocessing from~\cite{uda} to crop the images around the horses and get the training and validation sets. We also create a filtered \textbf{$\text{TDH}_{99}$} 99-images subset. To train YOLOv5s we also use the images containing zebras belonging to the COCO~\cite{cocodataset} dataset. We used this in two variants \textbf{$\text{CZ}_{1920}$} and \textbf{$\text{CZ}_{640}$}, signifying a different scaling factor of $1920\times1920$ and $640\times640$ respectively.

\noindent\textbf{Real-world data --- Aerial:}
The aerial datasets are called \textbf{RP} and \textbf{R123}. R123 consists of 104K images from three different experiments in which two DJI Mavic Pro were used to observe one to five adult individuals of the Grévy's zebra species. Each video is recorded at $3840\times2160@29.97\text{fps}$. We then manually temporally align and split the videos into subsequences of various lengths, in which at least one individual is always observed by the two UAVs simultaneously. Each zebra has then been enclosed in a manually drawn precise bounding box. We release these videos and the annotations in the context of this work. Finally, RP is a mix of images taken during the same experiments from both the UAVs (not necessarily belonging to R123) and low viewpoints, thanks to the use of some GoPros placed around the same arena containing the zebras. 

\subsection{Evaluation metrics}
The detection models are evaluated using the \textbf{mAP} and \textbf{mAP50}, i.e. the COCO standard average precision metric averaged over different IoU thresholds (mAP@[.5, .95]) and the PASCAL VOC’s metric (mAP@.5). The pose estimation models are instead evaluated using the percentage of correct keypoints (PCK). The threshold for the PCK is set to 5\% and 10\% of the maximum size of the bounding box. We indicate those respectively \textbf{$\text{P}_{0.05}$} and \textbf{$\text{P}_{0.1}$}. 

\subsection{Detection performance}
\label{sec:detection_results}
We train a YOLOv5s detector from scratch using randomly initialized weights and the same training protocol as in~\cite{bonetto-syn-zebras} for fairness. Specifically, training is conducted for 300 epochs with the default learning rate, using an input resolution of $1920\times1920$ pixels. This, except for the $\text{CZ}_{640}$ variant, which is trained at $640\times640$. Although it is common practice to test models on images resized to the training resolution, we evaluate all models at both $1920\times1920$ and $640\times640$.
Due to limited image availability, we utilize the entire $\text{A36}_\text{OZ}$ (1,200 images) and $\text{A10}_\text{OZ}$ (200 images) datasets for evaluation. During these tests, we also observed that at least nine images are shared between $\text{A10}_\text{OZ}$ and CZ.

\begin{figure*}[!ht]
    \centering
    
    \includegraphics[width=0.23\textwidth]{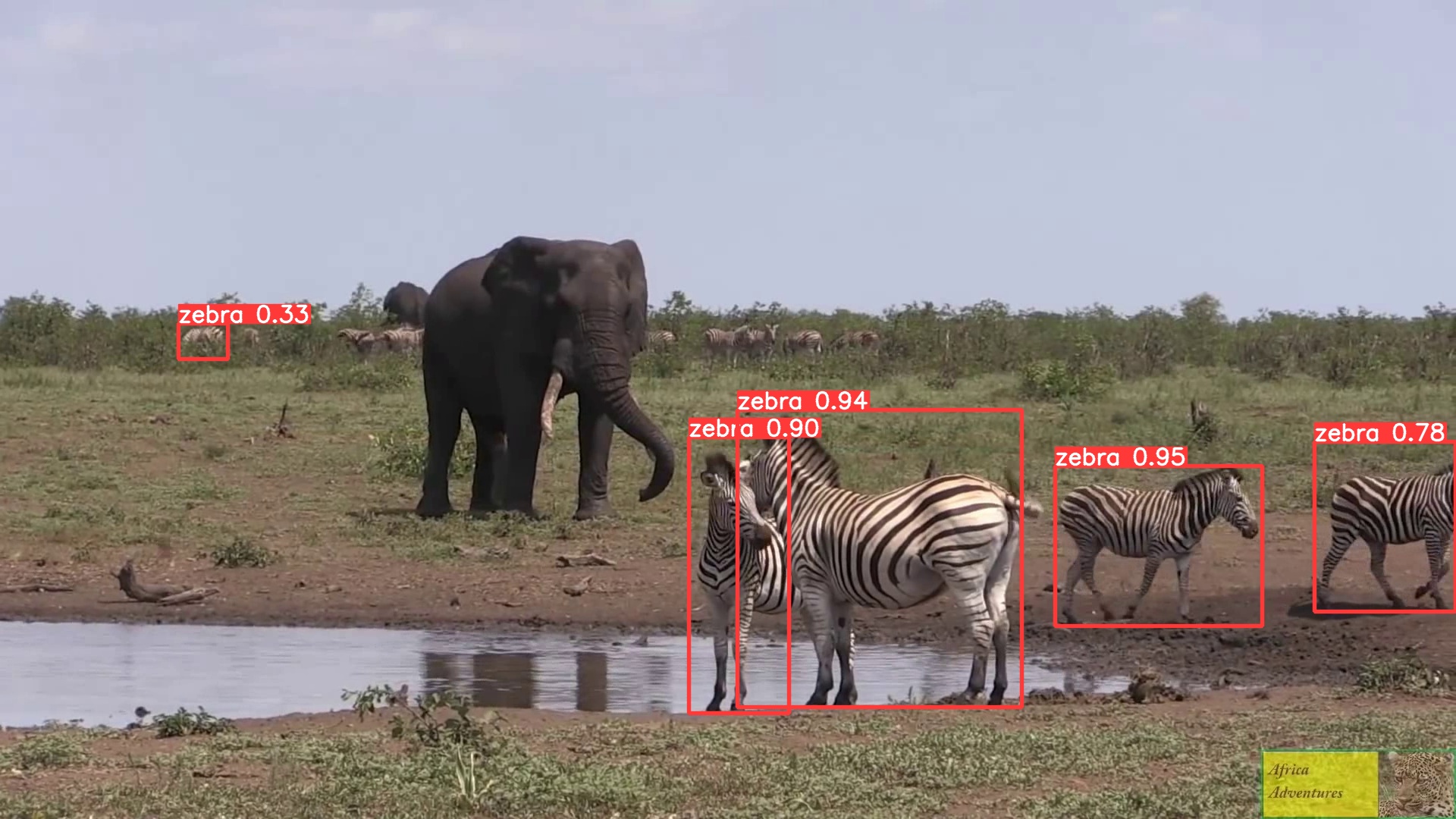}\hfill
    \includegraphics[width=0.23\textwidth]{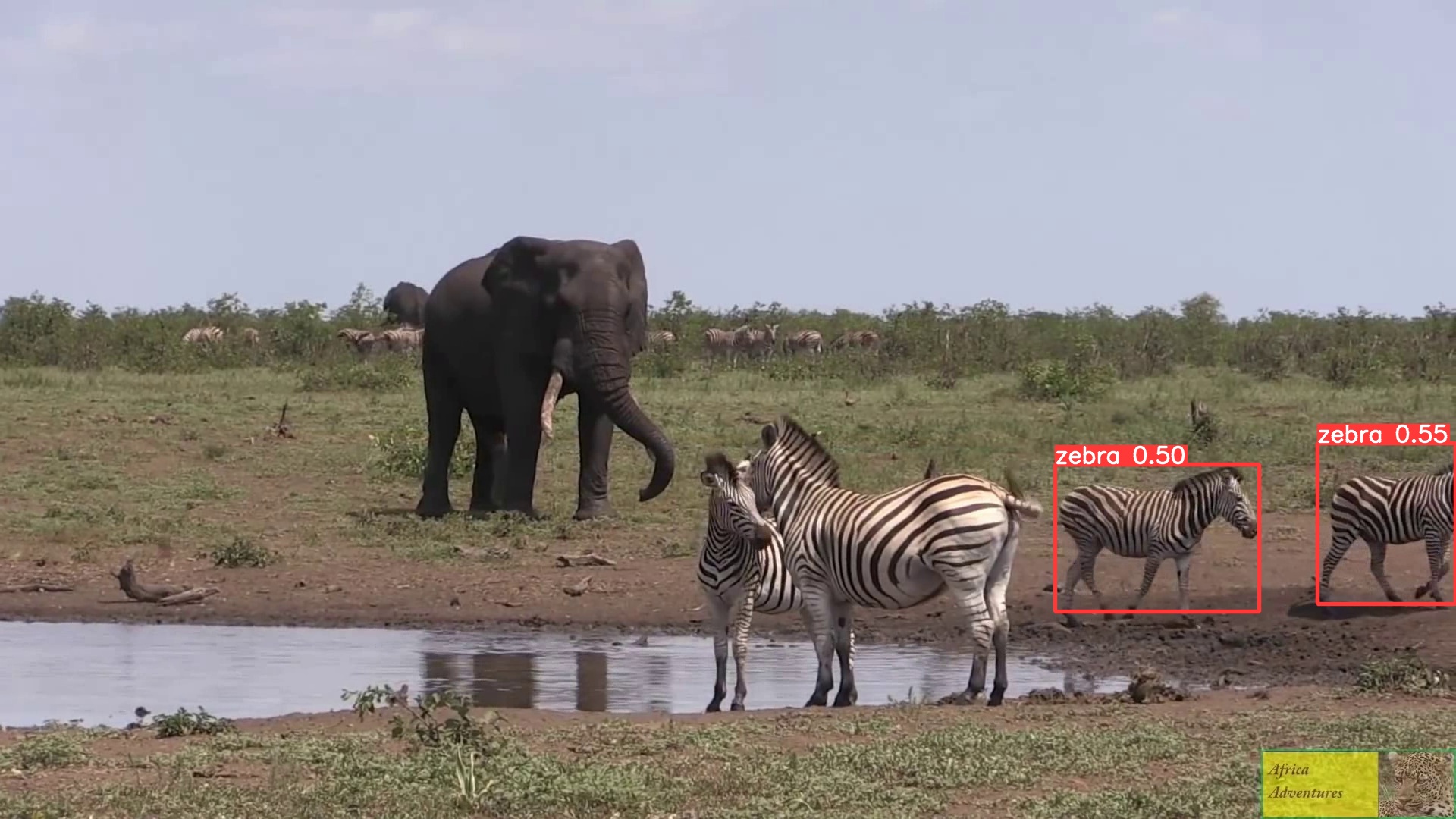}\hfill
    \includegraphics[width=0.23\textwidth]{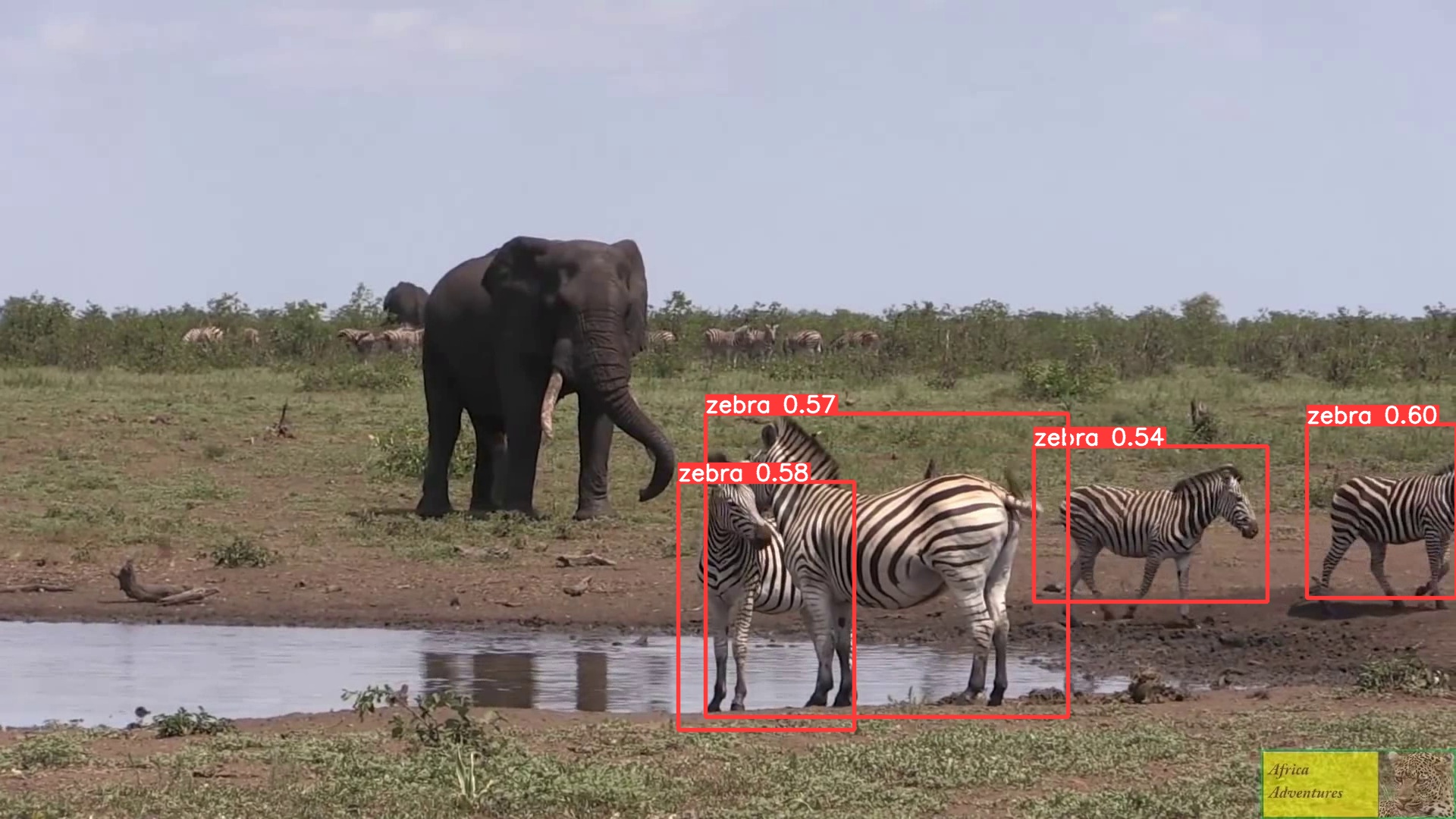}\hfill
    \includegraphics[width=0.23\textwidth]{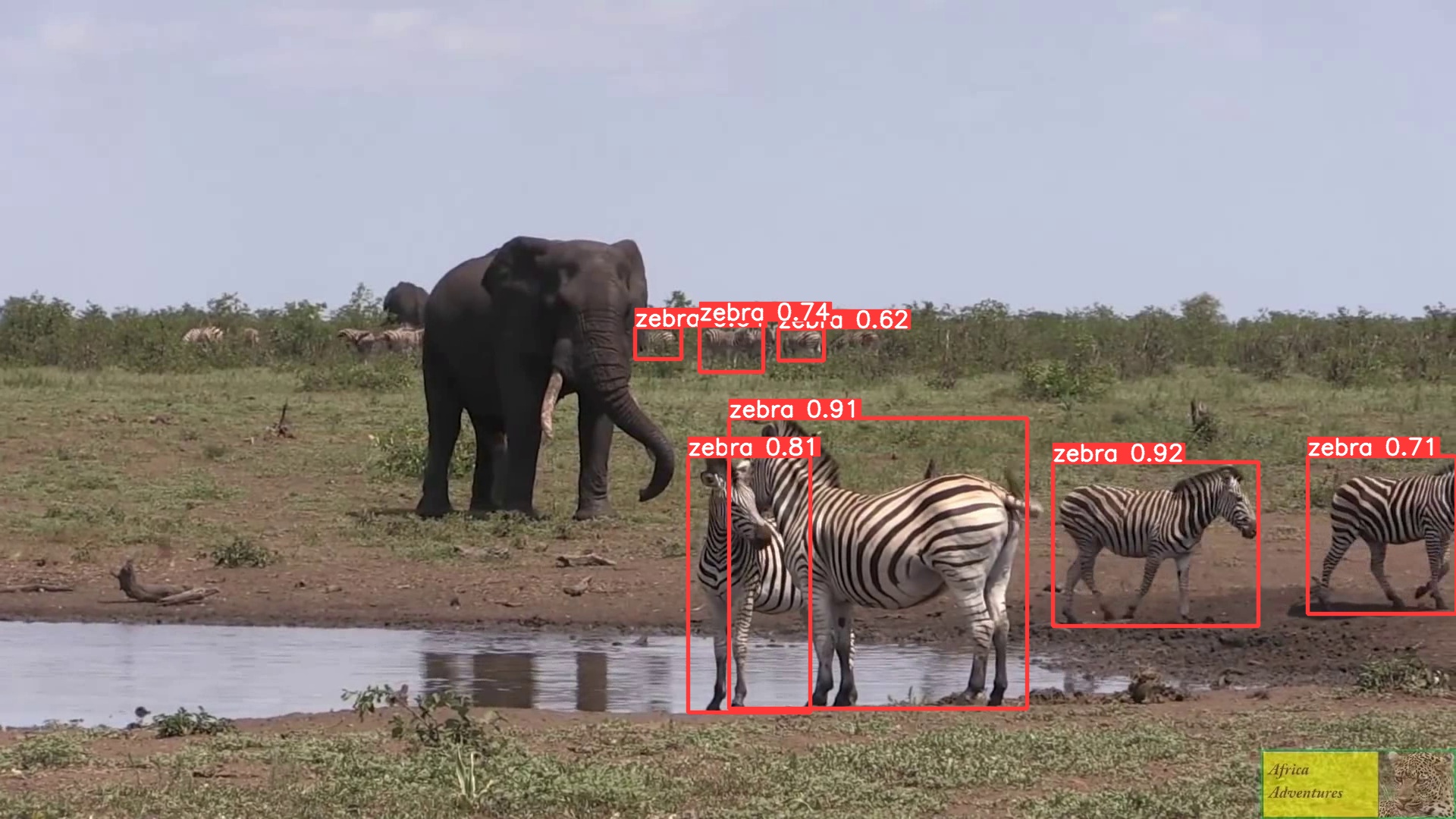}\hfill\\

    \subcaptionbox{Trained with $\text{CZ}_{1920}$}{\includegraphics[width=0.23\textwidth]{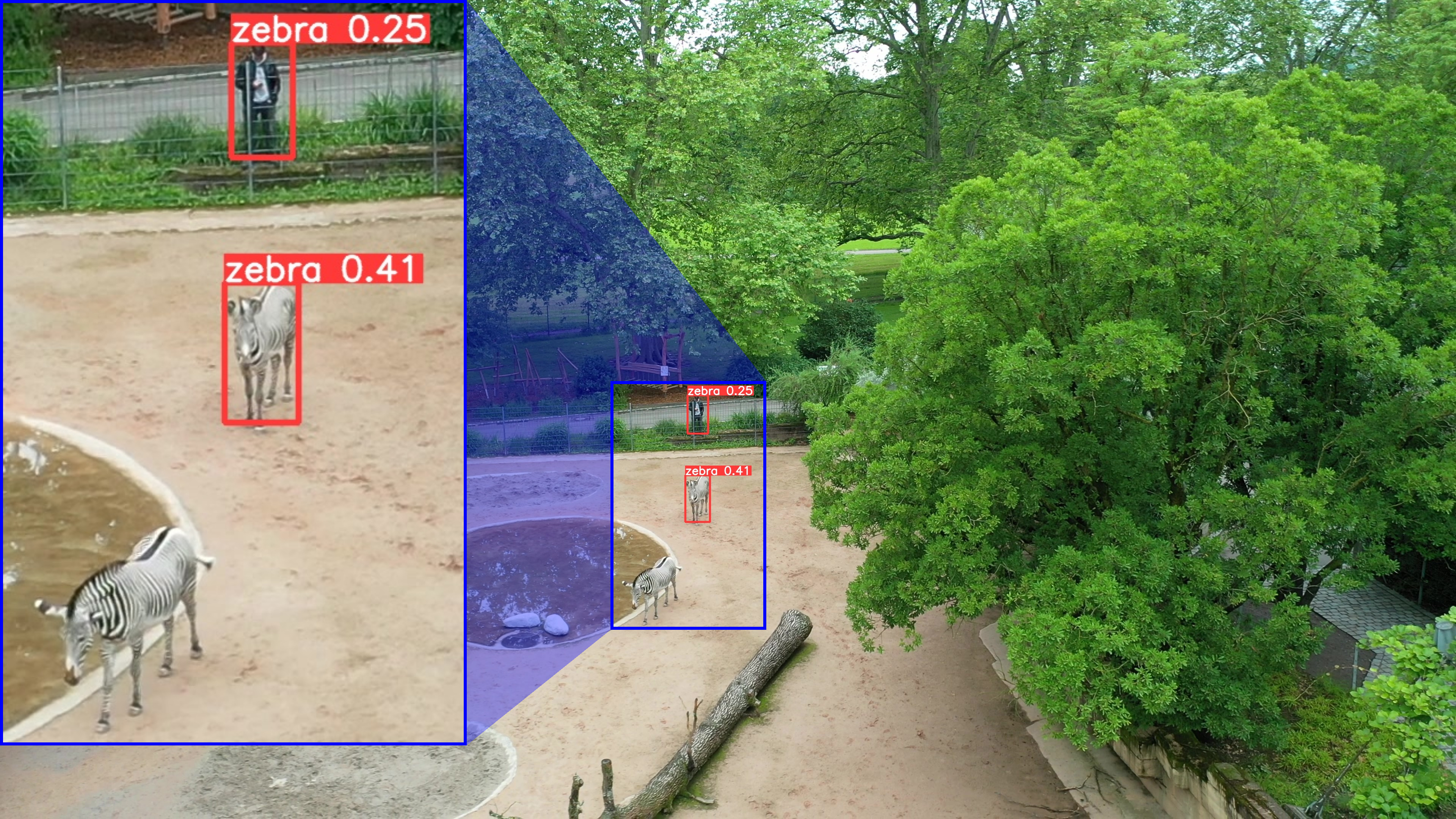}}\hfill
    \subcaptionbox{Trained with SC}{\includegraphics[width=0.23\textwidth]{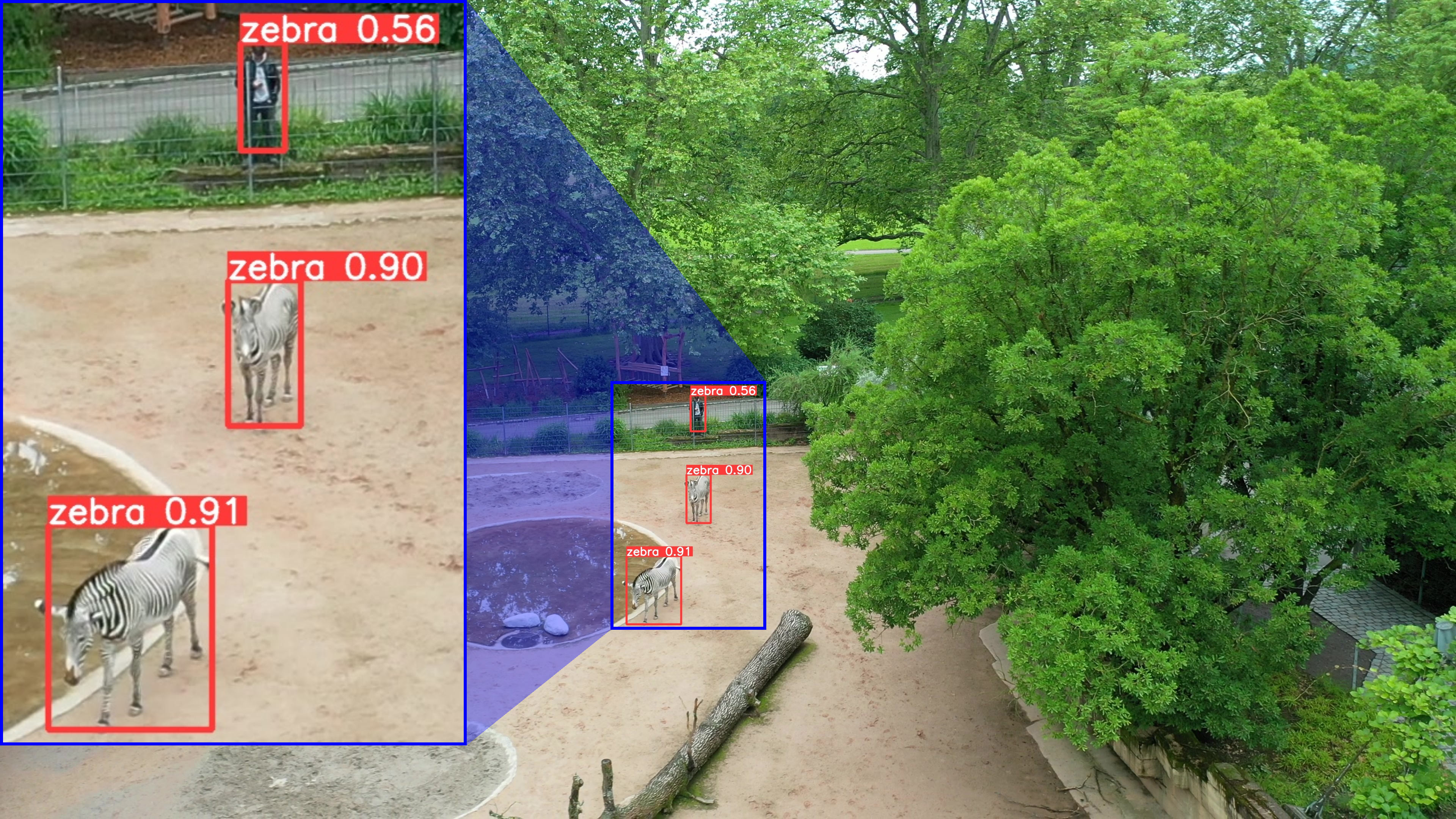}}\hfill     
    \subcaptionbox{Trained with $\text{SC}_\text{5K}$}{\includegraphics[width=0.23\textwidth]{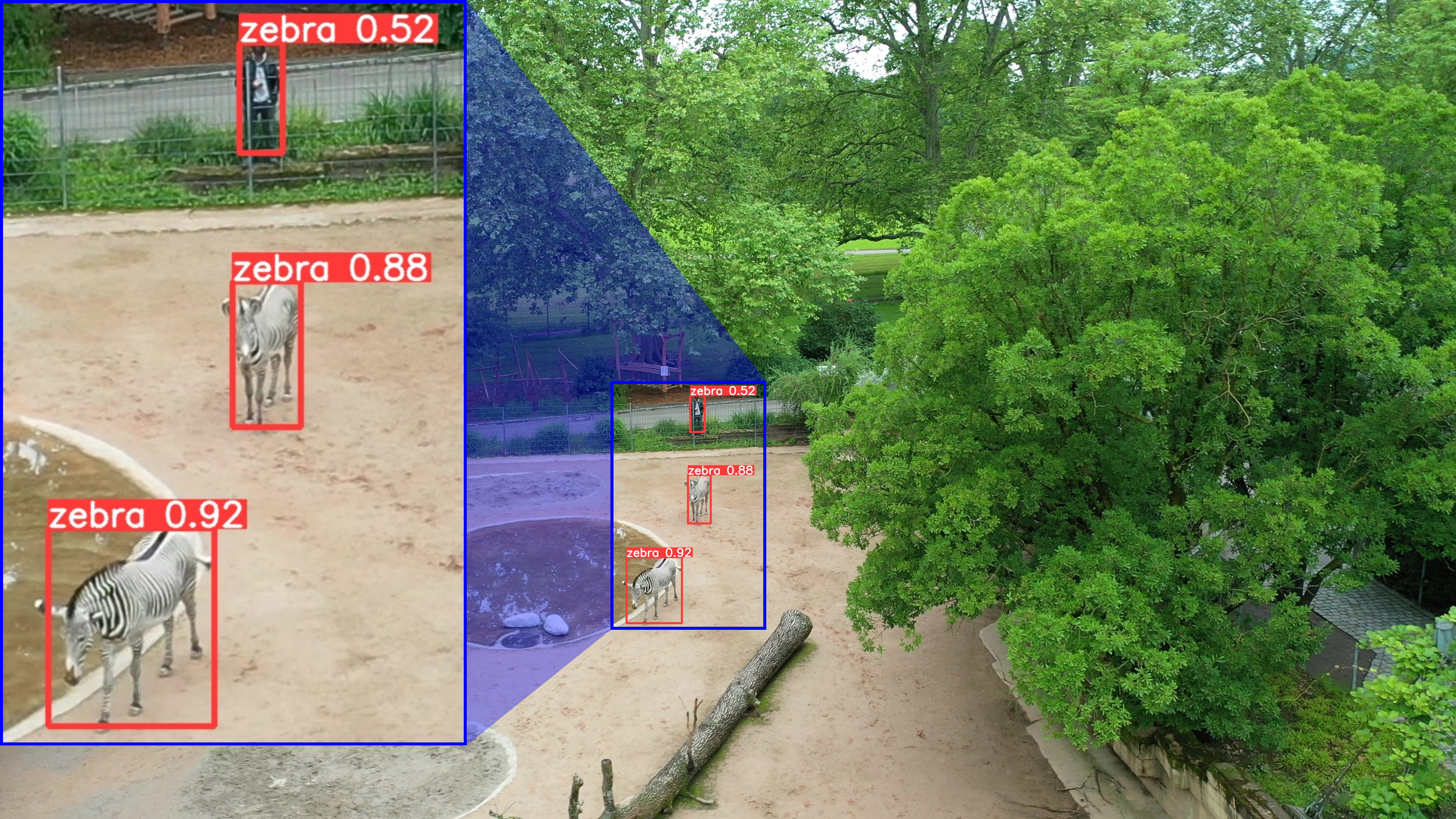}}\hfill
    \subcaptionbox{Trained with $\text{SC}_\text{5K}$+CZ$_{1920}$+RP}{\includegraphics[width=0.23\textwidth]{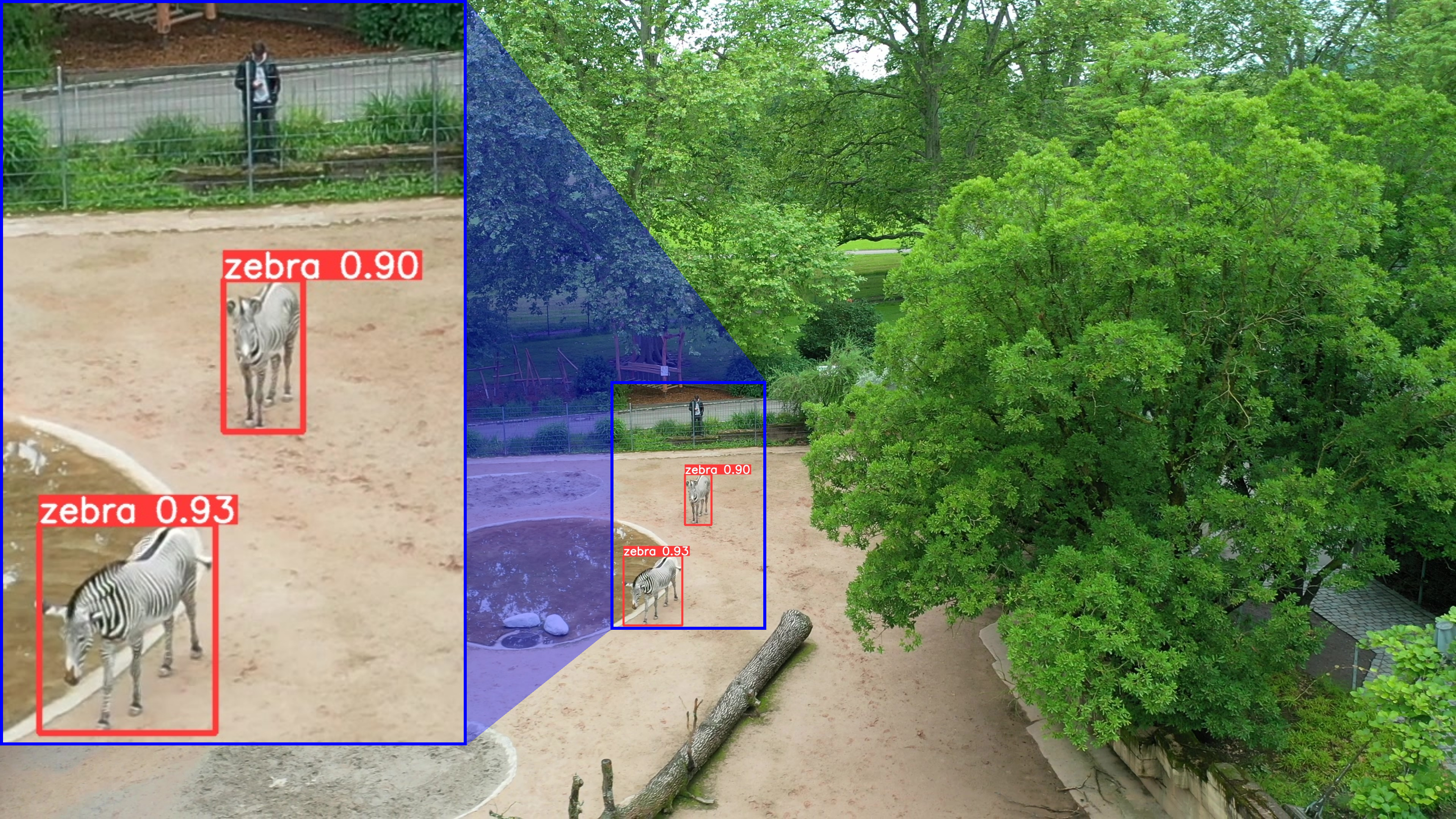}}\hfill

    \caption{YOLOv5s results on images from the APT-36K (top row) and our R123 (bottom row) datasets, using $1920\times1920$ resolution. }
    \label{fig:qualitative-yolo}
\end{figure*}

In~\cref{tab:yolo-1920}, we present the results obtained using images scaled to $1920\times1920$. Models trained with the new synthetic dataset, $\text{SC}_{\text{5K}}$, consistently outperform those trained on the previous SC dataset. For instance, on $\text{A36}_{\text{OZ}}$, training with $\text{SC}_{\text{5K}}$ yields a +36\% improvement in mAP50 compared to SC. On average, when YOLOv5s is trained solely on $\text{SC}_{\text{5K}}$, its performance matches or exceeds that of models trained exclusively on $\text{CZ}_{640}$, except on RP. Furthermore, the same model significantly outperforms $\text{CZ}_{1920}$ in both RP and R123, particularly on aerial images. The results obtained using images scaled to $640\times640$, shown in~\cref{tab:yolo-640}, reinforce these findings. Models trained with synthetic data continue to perform well on average, while those trained solely on COCO struggle on R123 and RP, further highlighting the versatility of our synthetic data.

\begin{table*}[!ht]
\centering
\resizebox{.85\textwidth}{!}{%
\begin{tabular}{l|cc|cc|cc|cc|cc|cc|cc}
    \toprule[1.5pt]
 & \multicolumn{8}{c|}{Only Zebras} & \multicolumn{4}{c|}{Various animals} & \multicolumn{2}{c}{Horses} \\ \cline{2-15} 
 \multicolumn{1}{r|}{Val. Set $\rightarrow$} & \multicolumn{2}{c|}{Zebra-300} & \multicolumn{2}{c|}{$\text{A10}_{\text{OZ}}$} & \multicolumn{2}{c|}{$\text{A36}_{\text{OZ}}$ } & \multicolumn{2}{c|}{Zebra-Zoo} & \multicolumn{2}{c|}{A10} & \multicolumn{2}{c|}{A36} & \multicolumn{2}{c}{TDH} \\ \cline{2-15} 
 Train Set $\downarrow$ & $\text{P}_{0.05}$ & $\text{P}_{0.1}$ & $\text{P}_{0.05}$ & $\text{P}_{0.1}$ & $\text{P}_{0.05}$ & $\text{P}_{0.1}$ & $\text{P}_{0.05}$ & $\text{P}_{0.1}$ & $\text{P}_{0.05}$ & $\text{P}_{0.1}$ & $\text{P}_{0.05}$ & $\text{P}_{0.1}$ & $\text{P}_{0.05}$ & $\text{P}_{0.1}$ \\ \hline
$\text{A10}_{\text{OZ}}$ & 0.372 & 0.621 & 0.353 & 0.569 & 0.252 & 0.507 & 0.353 & 0.625 & 0.084 & 0.191 & 0.079 & 0.180 & 0.288 & 0.521 \\
$\text{A36}_{\text{OZ}}$ & 0.378 & 0.608 & 0.358 & 0.552 & 0.328 & 0.544 & 0.251 & 0.476 & 0.055 & 0.133 & 0.062 & 0.143 & 0.165 & 0.345 \\
SpacNet & 0.207 & 0.411 & 0.275 & 0.465 & 0.245 & 0.419 & 0.139 & 0.327 & 0.050 & 0.122 & 0.042 & 0.106 & 0.086 & 0.227 \\
A10 & 0.685 & 0.889 & 0.629 & 0.823 & 0.532 & 0.768 & 0.664 & 0.876 & 0.516 & 0.718 & 0.465 & 0.661 & 0.596 & 0.853 \\
A36 & 0.691 & 0.883 & 0.602 & 0.783 & 0.585 & 0.777 & 0.495 & 0.774 & 0.366 & 0.557 & 0.531 & 0.722 & 0.638 & 0.875 \\
TDH & 0.080 & 0.192 & 0.059 & 0.158 & 0.039 & 0.126 & 0.050 & 0.160 & 0.057 & 0.141 & 0.064 & 0.156 & 0.949 & 0.974 \\
SC & 0.466 & 0.686 & 0.292 & 0.511 & 0.273 & 0.447 & 0.395 & 0.600 & 0.082 & 0.178 & 0.095 & 0.200 & 0.294 & 0.475 \\
\textit{---filtered} & 0.608 & 0.790 & 0.345 & 0.573 & 0.340 & 0.508 & 0.522 & 0.707 & 0.090 & 0.184 & 0.106 & 0.207 & 0.324 & 0.505 \\ \hline
SC+$\text{A10}_{\text{OZ}}$ & 0.847 & 0.967 & 0.675 & 0.879 & 0.705 & 0.878 & 0.819 & 0.945 & 0.165 & 0.303 & 0.169 & 0.308 & 0.434 & 0.659 \\
 \textit{---filtered}& 0.864 & 0.966 & 0.657 & 0.869 & 0.716 & 0.867 & 0.796 & 0.932 & 0.174 & 0.307 & 0.180 & 0.315 & 0.473 & 0.681 \\
SC+$\text{A36}_{\text{OZ}}$ & 0.844 & 0.961 & 0.657 & 0.848 & 0.768 & 0.918 & 0.855 & 0.970 & 0.190 & 0.342 & 0.243 & 0.410 & 0.546 & 0.777 \\
\textit{---filtered}& 0.858 & 0.956 & 0.651 & 0.831 & 0.779 & 0.920 & 0.851 & 0.963 & 0.198 & 0.341 & 0.257 & 0.415 & 0.568 & 0.782 \\
SC+$\text{A10}_{\text{99}}$ & 0.750 & 0.938 & 0.585 & 0.803 & 0.605 & 0.832 & 0.737 & 0.928 & 0.119 & 0.231 & 0.131 & 0.252 & 0.410 & 0.626 \\
 \textit{---filtered}& 0.828 & 0.950 & 0.591 & 0.809 & 0.636 & 0.824 & 0.750 & 0.918 & 0.129 & 0.236 & 0.144 & 0.259 & 0.443 & 0.644 \\
SpacNet+$\text{A10}_{\text{99}}$ & 0.525 & 0.777 & 0.462 & 0.674 & 0.437 & 0.676 & 0.482 & 0.768 & 0.091 & 0.200 & 0.080 & 0.179 & 0.199 & 0.408 \\
SC+$\text{TDH}_{\text{99}}$ & 0.490 & 0.707 & 0.321 & 0.528 & 0.336 & 0.538 & 0.396 & 0.580 & 0.128 & 0.260 & 0.155 & 0.303 & 0.658 & 0.873 \\
 \textit{---filtered}& 0.634 & 0.813 & 0.373 & 0.591 & 0.421 & 0.603 & 0.533 & 0.688 & 0.146 & 0.274 & 0.177 & 0.321 & 0.685 & 0.872 \\ \hline
SC+A10 & \textit{0.891} & \textit{0.977} & \textbf{0.714} & \textbf{0.894} & 0.756 & 0.911 & 0.882 & 0.977 & \textit{0.643} & \textit{0.825} & 0.626 & 0.798 & 0.717 & 0.918 \\
 \textit{---filtered}& \textbf{0.898} & \textit{0.973} & \textit{0.699} & \textit{0.889} & 0.777 & 0.909 & 0.881 & 0.971 & \textbf{0.673} & \textbf{0.829} & 0.650 & 0.806 & 0.754 & 0.912 \\
SC+A36 & 0.861 & 0.965 & 0.697 & 0.867 & \textit{0.787} & \textit{0.936} & \textbf{0.908} & \textbf{0.985} & 0.522 & 0.709 & \textit{0.678} & \textit{0.840} & 0.729 & 0.923 \\
 \textit{---filtered}& 0.868 & 0.960 & 0.696 & 0.867 & \textbf{0.797} & \textbf{0.938} & \textit{0.900} & \textit{0.983} & 0.552 & 0.719 & \textbf{0.703} & \textbf{0.849} & 0.763 & 0.916 \\
SC+TDH & 0.524 & 0.754 & 0.374 & 0.595 & 0.334 & 0.555 & 0.465 & 0.714 & 0.146 & 0.295 & 0.189 & 0.379 & \textbf{0.969} & \textbf{0.989} \\
\textit{---filtered} & 0.668 & 0.838 & 0.432 & 0.657 & 0.402 & 0.606 & 0.588 & 0.807 & 0.164 & 0.307 & 0.208 & 0.394 & \textit{0.965} & \textit{0.986} \\
    \bottomrule[1.5pt]
\end{tabular}%
}
\caption{PCK@0.05 and PCK@0.1 of the ViTPose+ models trained on the specified dataset (row) with all \textbf{weights randomly initialized} and evaluated on the specified data (columns). We put in bold the best results, and in italics the second best.}
\label{tab:vit-nopt}
\end{table*}

Notably, the evaluation resolution has a significant impact on performance. This can be observed in the differences between the results of $\text{CZ}_{1920}$ and $\text{CZ}_{640}$ when evaluated at their corresponding or different validation image sizes. For instance, when tested on $1920\times1920$ images, $\text{CZ}_{640}$ outperforms $\text{CZ}_{1920}$ on RP and R123, while $\text{CZ}_{1920}$ achieves better results on $\text{A36}_{\text{OZ}}$ and $\text{A10}_{\text{OZ}}$. Additionally, performance on aerial images (R123, RP) degrades when the input is downscaled to $640\times640$ pixels, whereas models perform better on `common' viewpoints under the same conditions (and vice versa). This is likely due to the relative size of individuals observed during training and validation. Specifically, large images, such as those from R123, when resized to $640\times640$, further reduce the size of objects, making detection more challenging—especially for models trained predominantly on high-resolution data (i.e. all except $\text{CZ}_{640}$). Conversely, not upscaling smaller images like those in $\text{A36}{\text{OZ}}$ or $\text{A10}_{\text{OZ}}$ to $1920\times1920$ preserves the small object scale, which benefits models trained on larger images by a considerable margin. Therefore, there is no clear evidence that matching training and validation resolutions always leads to better performance; instead, the optimal resolution depends on the distribution of object sizes.

\begin{table*}[!th]
\centering
\resizebox{.85\textwidth}{!}{%
\begin{tabular}{l|cc|cc|cc|cc|cc|cc|cc}
    \toprule[1.5pt]
 & \multicolumn{8}{c|}{Zebras} & \multicolumn{4}{c|}{Various animals} & \multicolumn{2}{c}{Horses} \\ \cline{2-15} 
 \multicolumn{1}{r|}{Val. Set $\rightarrow$} & \multicolumn{2}{c|}{Zebra-300} & \multicolumn{2}{c|}{$\text{A10}_{\text{OZ}}$ } & \multicolumn{2}{c|}{$\text{A36}_{\text{OZ}}$ } & \multicolumn{2}{c|}{Zebra-Zoo} & \multicolumn{2}{c|}{A10} & \multicolumn{2}{c|}{A36} & \multicolumn{2}{c}{TDH} \\ \cline{2-15} 
 Train Set $\downarrow$ & $\text{P}_{0.05}$ & $\text{P}_{0.1}$ & $\text{P}_{0.05}$ & $\text{P}_{0.1}$ & $\text{P}_{0.05}$ & $\text{P}_{0.1}$ & $\text{P}_{0.05}$ & $\text{P}_{0.1}$ & $\text{P}_{0.05}$ & $\text{P}_{0.1}$ & $\text{P}_{0.05}$ & $\text{P}_{0.1}$ & $\text{P}_{0.05}$ & $\text{P}_{0.1}$ \\ \hline
 $\text{A10}_{\text{OZ}}$ & 0.753 & 0.902 & 0.673 & 0.815 & 0.654 & 0.833 & 0.740 & 0.887 & 0.264 & 0.453 & 0.243 & 0.414 & 0.463 & 0.742 \\
$\text{A36}_{\text{OZ}}$ & 0.849 & 0.954 & 0.690 & 0.833 & 0.778 & 0.923 & 0.865 & 0.962 & 0.230 & 0.410 & 0.271 & 0.458 & 0.422 & 0.738 \\
SpacNet & 0.533 & 0.792 & 0.441 & 0.644 & 0.434 & 0.655 & 0.419 & 0.724 & 0.111 & 0.235 & 0.092 & 0.202 & 0.198 & 0.414 \\
A10 & 0.888 & 0.979 & \textbf{0.779} & \textit{0.900} & 0.789 & 0.918 & 0.892 & 0.980 & \textbf{0.710} & \textbf{0.868} & 0.672 & 0.822 & 0.707 & 0.908 \\
A36 & 0.880 & 0.968 & \textit{0.747} & 0.870 & 0.805 & 0.936 & 0.902 & 0.984 & 0.579 & 0.759 & 0.712 & 0.867 & 0.724 & 0.924 \\
TDH & 0.145 & 0.291 & 0.117 & 0.273 & 0.174 & 0.337 & 0.086 & 0.223 & 0.138 & 0.281 & 0.164 & 0.330 & \textit{0.965} & \textit{0.986} \\
SC & 0.527 & 0.736 & 0.363 & 0.569 & 0.324 & 0.514 & 0.525 & 0.730 & 0.083 & 0.177 & 0.099 & 0.204 & 0.276 & 0.450 \\
 \textit{---filtered}& 0.689 & 0.859 & 0.434 & 0.647 & 0.409 & 0.588 & 0.720 & 0.875 & 0.093 & 0.183 & 0.112 & 0.212 & 0.304 & 0.475 \\ \hline
SC+$\text{A10}_{\text{OZ}}$ & 0.864 & 0.981 & 0.671 & 0.866 & 0.742 & 0.905 & 0.905 & 0.979 & 0.151 & 0.279 & 0.160 & 0.285 & 0.359 & 0.559 \\
 \textit{---filtered}& 0.899 & 0.981 & 0.665 & 0.865 & 0.771 & 0.906 & 0.911 & 0.974 & 0.162 & 0.282 & 0.173 & 0.291 & 0.397 & 0.581 \\
SC+$\text{A36}_{\text{OZ}}$ & 0.878 & 0.972 & 0.687 & 0.874 & 0.806 & 0.947 & 0.935 & 0.990 & 0.215 & 0.372 & 0.275 & 0.447 & 0.534 & 0.783 \\
\textit{---filtered}& 0.891 & 0.970 & 0.675 & 0.862 & \textit{0.823} & 0.950 & \textit{0.942} & 0.988 & 0.221 & 0.369 & 0.288 & 0.451 & 0.556 & 0.781 \\
SC+$\text{A10}_{\text{99}}$ & 0.757 & 0.941 & 0.600 & 0.828 & 0.629 & 0.844 & 0.813 & 0.957 & 0.119 & 0.234 & 0.136 & 0.258 & 0.343 & 0.536 \\
\textit{---filtered} & 0.863 & 0.960 & 0.621 & 0.830 & 0.677 & 0.856 & 0.876 & 0.962 & 0.128 & 0.238 & 0.150 & 0.265 & 0.378 & 0.560 \\
SpacNet+$\text{A10}_{\text{99}}$ & 0.796 & 0.931 & 0.636 & 0.778 & 0.624 & 0.814 & 0.743 & 0.925 & 0.190 & 0.353 & 0.149 & 0.296 & 0.289 & 0.565 \\
SC+$\text{TDH}_{\text{99}}$ & 0.564 & 0.773 & 0.385 & 0.603 & 0.345 & 0.545 & 0.547 & 0.731 & 0.136 & 0.268 & 0.172 & 0.324 & 0.678 & 0.884 \\
 \textit{---filtered}& 0.745 & 0.901 & 0.456 & 0.696 & 0.440 & 0.624 & 0.731 & 0.875 & 0.154 & 0.283 & 0.195 & 0.339 & 0.707 & 0.878 \\ \hline
SC+A10 & \textit{0.906} & \textbf{0.984} & 0.728 & \textit{0.900} & 0.782 & 0.916 & 0.932 & \textbf{0.993} & 0.670 & 0.842 & 0.657 & 0.819 & 0.732 & 0.922 \\
 \textit{---filtered}& \textbf{0.916} & \textit{0.982} & 0.715 & \textbf{0.901} & 0.817 & 0.922 & \textbf{0.944} & \textit{0.992} & \textit{0.703} & \textit{0.849} & 0.684 & 0.829 & 0.774 & 0.917 \\
SC+A36 & 0.890 & 0.976 & 0.724 & 0.897 & 0.820 & \textit{0.953} & 0.934 & \textbf{0.993} & 0.575 & 0.757 & \textit{0.719} & \textit{0.871} & 0.760 & 0.936 \\
 \textit{---filtered}& 0.905 & 0.973 & 0.707 & 0.893 & \textbf{0.845} & \textbf{0.958} & 0.938 & \textit{0.992} & 0.610 & 0.771 & \textbf{0.750} & \textbf{0.882} & 0.800 & 0.934 \\
SC+TDH & 0.586 & 0.800 & 0.412 & 0.624 & 0.384 & 0.598 & 0.606 & 0.801 & 0.166 & 0.321 & 0.221 & 0.417 & \textbf{0.967} & \textbf{0.989} \\
 \textit{---filtered}& 0.754 & 0.902 & 0.489 & 0.710 & 0.477 & 0.677 & 0.793 & 0.920 & 0.190 & 0.338 & 0.242 & 0.435 & 0.963 & \textit{0.986} \\
    \bottomrule[1.5pt]
\end{tabular}%
}
\caption{PCK@0.05 and PCK@0.1 of the ViTPose+ models trained on the specified dataset (row) using a \textbf{MAE pre-trained backbone} and evaluated on the specified data (columns). We put in bold the best results, and in italics the second best.}\label{tab:vit-pt}
\end{table*}

The best-performing models are those trained with a combination of RP, $\text{CZ}_{1920}$, and $\text{SC}_{\text{5K}}$. However, no single model consistently achieves strong performance across all datasets and scaling sizes. Nonetheless, the results demonstrate that synthetic data alone can yield competitive performance, often matching or surpassing models trained solely on real data in both average and weighted mAP50 and mAP scores. The significant performance drop on $\text{A10}_{\text{OZ}}$ compared to $\text{A36}_{\text{OZ}}$ (8–25\%) suggests that detection performance cannot be assumed even when using models trained on real data, further highlighting the importance of comprehensive testing and diverse training data.

Finally, the qualitative comparison in~\cref{fig:qualitative-yolo} (and Supp.~\cref{fig:qualitative-yolo-supp}) illustrates how the $\text{SC}_\text{5K}$ model improves over the standard SC on the APT-36K dataset (top row, column (c)). In contrast, the baseline model struggles with aerial images (central and bottom rows, column (a)). As expected, the mixed model (last column) demonstrates the best overall performance. Additionally, when compared with~\cref{fig:img_compare}, we observe that zebras detected in the background of the APT-36K image do not contribute to accuracy due to mislabeling, whereas the correctly not labeled elephant lowers the computed accuracy, affecting the final results. This suggests that incorporating real-world data as validation for synthetic-based training could further enhance performance.

\subsection{2D pose estimation performance}
\label{sec:pose_results}

We employ the popular ViTPose architecture in its \mbox{ViTPose+}~\cite{vitposeplus} variation, following standard training settings: 210 epochs with decay steps at 170 and 200, Adam optimizer with a $5\mathrm{e}{-4}$ learning rate, and Gaussian heatmaps as the target type. Ground truth bounding boxes are used during both training and validation. We train the \textit{small} network model using 17 keypoints for real datasets and 27 keypoints for synthetic or mixed datasets. Additionally, due to minor annotation misalignments in our synthetic data (specifically, the thighs and tail base differing by a few pixels), we also report a \textit{filtered} average result that excludes these keypoints. Training is conducted both from scratch and with a default MAE pre-trained backbone.

First, in~\cref{tab:vit-nopt}, we analyze results obtained without pre-training. The model trained only on SC achieves performance comparable to those trained solely on zebra datasets ($\text{A36}_{\text{OZ}}$ or $\text{A10}_{\text{OZ}}$) for both PCK metrics. However, it demonstrates better generalization on horses (TDH) and across the full A10 and A36 datasets, with improvements exceeding 13\%. A similar trend is observed when comparing SC to SpacNet, with gains ranging from 3\% to 25\%. Moreover, the \textit{filtered} SC model significantly outperforms those trained on $\text{A36}_{\text{OZ}}$ and $\text{A10}_{\text{OZ}}$. Despite SC’s limited pose diversity due to the constrained number of animated frames (\cref{sec:approach}), it effectively trains a 2D pose estimator for real images. As expected, training on the full A10 and A36 datasets achieves the highest performance among single-dataset models, benefiting from domain transfer between similar species and diverse animal instances. However, when comparing Zebra-300 and Zebra-Zoo, the \textit{filtered} SC model performs similarly or better than models trained on A10 and A36. Additionally, SC generalizes better to TDH horses than TDH data does to zebras. Integrating real-world data further enhances performance, even with minimal additions. For instance, SC+$\text{A}10_{99}$ significantly outperforms SpacNet+$\text{A}10_{99}$ in both zebra keypoint detection and generalization, with PCK improvements of up to 20\%. Similarly, SC+$\text{TDH}_{99}$ achieves TDH performance comparable to models trained on the full A10 or A36 datasets. Finally, combining SC with the entirety of A10 or A36 yields minimal gains, except for improvements on non-zebra datasets.

In~\cref{tab:vit-pt}, we show how using a pre-trained backbone generally enhances performance across all models, datasets, and metrics by leveraging real-world information. Training with $\text{A36}_{\text{OZ}}$ or $\text{A10}_{\text{OZ}}$ achieves PCK@0.05 values above 60\% on all zebra-related validation sets. Notably, $\text{A36}_{\text{OZ}}$ consistently outperforms $\text{A10}_{\text{OZ}}$ on those sets but exhibits poorer generalization to the full A10, A36, and TDH datasets. Using SpacNet data results in a notable drop in performance, with PCK@0.05 ranging from 42\% to 53\% and PCK@0.1 between 64\% and 79\%. In contrast, training with SC data achieves similar results to SpacNet, despite not employing any augmentation or style transfer techniques. The SC\textit{-filtered} results, which account for keypoint misalignment, significantly improve PCK@0.05 in the Zebra-300 and Zebra-Zoo datasets by approximately 15–31\% over SpacNet, demonstrating the superiority of our synthetic data. While SC\textit{-filtered} performance is still about 15\% lower than $\text{A36}_{\text{OZ}}$, it closely matches $\text{A10}_{\text{OZ}}$, both of which contain only real zebras. Similar to the non-pre-trained case, SC data exhibits strong generalization on horses, achieving a 48\% PCK@0.1. Performance improves further as real-world data is incorporated, with mixed datasets consistently outperforming single ones. Interestingly, when comparing pre-trained and non-pre-trained models (\cref{tab:vit-pt} vs.~\cref{tab:vit-nopt}), we observe that SC alone reduces the need for pre-training. Specifically, training with just SC+$\text{A}10_{99}$ (i.e. our synthetic dataset plus 99 real images from A10) achieves results comparable to the best model trained with a pre-trained backbone.

\begin{figure}[ht]
    \centering
    \includegraphics[width=0.49\linewidth]{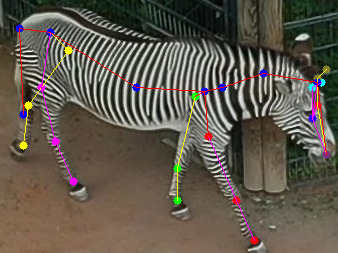}\hfill
    \includegraphics[scale=0.34]{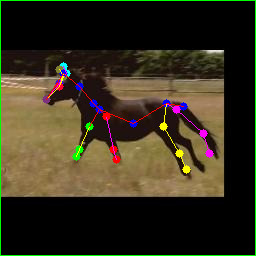}\hfill
    \caption{Examples of ViTPose+ estimations when trained using randomly initialized weights and only on synthetic data.}
    \label{fig:vit-pose-main}
\end{figure}

Qualitative analysis shows how models trained on synthetic data, e.g.~\cref{fig:vit-pose-main}, achieve good overall performance. Moreover, we can notice that most errors stem from swapped limbs or incorrect leg associations (see Supp.~\cref{fig:vit_drone_scr-supp,fig:vit_drone_pre-supp}). These errors can be attributed to differences in data generation and annotation practices. In our synthetic dataset, keypoints are extracted directly from the surface of the animal, and bounding boxes are pixel-tight. In contrast, human annotators typically place keypoints where they estimate the limb to be, and bounding boxes may not be tightly cropped (\cref{fig:img_compare}). Additionally, synthetic zebras sometimes intersect with the terrain, complicating the learning of hoof positions. Moreover, these can be linked to the limited pose diversity in the synthetic model and to frequent occurrences of small, far-away zebras, which could make some keypoints ambiguous. Finally, our use of instance masks might cause some keypoints to be labeled as ``visible'' in the training data, even if occluded, impacting our performance by making the model try to always predict something. Interestingly, one of the best-performing models, which was trained on A10 with a pre-trained backbone, still struggles with keypoint estimation in aerial views. In contrast, the model trained on synthetic data without pre-training performs well across all real-world images despite not using anatomically perfect keypoints. This is probably why we can observe slightly better performance of SC+$\text{A10}_{99}$ over SC+$\text{A10}$ in our Supp. video, as 99 images might be enough to ground the model, but will still keep a stronger aerial-related prior. Finally, the primary challenge in generalizing to horses can be identified in their uniform coloring, which lacks distinctive features. Additionally, unusual cropping confuses the model in determining the correct side of the animal and keypoint placement. Additional qualitative results can be found in the Supp. Mat. (\cref{fig:vit_drone_scr-supp,fig:vit_drone_pre-supp,fig:vit_zoo_no-pre-supp,fig:vit_zoo_pre-supp,fig:vit_horse_scra-supp,fig:vit_horse_pre-supp}), comparing datasets and training methods.
		\section{Conclusions}
\label{sec:conc}

In this paper, we show that synthetic data alone can train object detection and 2D pose models without style transfer techniques or real-world supervision. Our method outperforms~\cite{bonetto-syn-zebras} across datasets and generalizes well to real zebra poses, even with limited asset variability. Our findings highlight the effectiveness of the generation strategy, which prioritizes visual realism and variability, in addressing these tasks without needing pre-trained networks or complex models. Our pipeline enables fast, scalable, and precise data creation, adapting easily to new conditions, supporting challenging scenarios like wildlife and aerial views, and can generate far more consistent data than real-world datasets, allowing rapid, otherwise impractical iterations. Through our extensive validation, we also highlight that: i) Matching YOLO training and test sizes is not always optimal; ii) Large-scale synthetic data can reduce or replace the need for real data and pre-trained models; iii) Robust evaluation requires diverse test data. In future work, we will extend this work by leveraging depth maps and parts-segmentation for finer visibility estimates, increasing texture and shape variability to extend the work to 3D pose estimation and individual identification, and will introduce other species.
		\vspace{-20pt}
		{
			\small
			\bibliographystyle{ieeenat_fullname}
			\bibliography{main.bbl}

\begin{thebibliography}{41}
\providecommand{\natexlab}[1]{#1}
\providecommand{\url}[1]{\texttt{#1}}
\expandafter\ifx\csname urlstyle\endcsname\relax
  \providecommand{\doi}[1]{doi: #1}\else
  \providecommand{\doi}{doi: \begingroup \urlstyle{rm}\Url}\fi

\bibitem[Beery et~al.(2020)Beery, Liu, Morris, Piavis, Kapoor, Joshi, Meister,
  and Perona]{beery2020synthetic}
Sara Beery, Yang Liu, Dan Morris, Jim Piavis, Ashish Kapoor, Neel Joshi, Markus
  Meister, and Pietro Perona.
\newblock Synthetic examples improve generalization for rare classes.
\newblock In \emph{Proceedings of the ieee/cvf winter conference on
  applications of computer vision}, pages 863--873, 2020.

\bibitem[Black et~al.(2023)Black, Patel, Tesch, and Yang]{bedlam}
Michael~J. Black, Priyanka Patel, Joachim Tesch, and Jinlong Yang.
\newblock {BEDLAM}: A synthetic dataset of bodies exhibiting detailed lifelike
  animated motion.
\newblock In \emph{Proceedings IEEE/CVF Conf.~on Computer Vision and Pattern
  Recognition (CVPR)}, pages 8726--8737, 2023.

\bibitem[Bonetto and Ahmad(2023)]{bonetto-syn-zebras}
Elia Bonetto and Aamir Ahmad.
\newblock Synthetic data-based detection of zebras in drone imagery.
\newblock In \emph{2023 European Conference on Mobile Robots (ECMR)}, pages
  1--8, 2023.

\bibitem[Bonetto et~al.(2023)Bonetto, Xu, and Ahmad]{GRADE}
Elia Bonetto, Chenghao Xu, and Aamir Ahmad.
\newblock {GRADE}: Generating realistic animated dynamic environments for
  robotics research.
\newblock \emph{arXiv preprint arXiv:2303.04466}, 2023.

\bibitem[Bray and Edwards(1999)]{bray1999body}
RE Bray and MS Edwards.
\newblock Body condition scoring of captive (zoo) equids.
\newblock In \emph{Proceedings of the Third Conference on Zoo and Wildlife
  Nutrifion}, 1999.

\bibitem[Breed et~al.(2019)Breed, Meyer, Steyl, Goddard, Burroughs, and
  Kohn]{Breed2019}
Dorothy Breed, Leith C~R Meyer, Johan C~A Steyl, Amelia Goddard, Richard
  Burroughs, and Tertius~A Kohn.
\newblock Conserving wildlife in a changing world: Understanding capture
  myopathy—a malignant outcome of stress during capture and translocation.
\newblock \emph{Conservation Physiology}, 7\penalty0 (1), 2019.

\bibitem[Cao et~al.(2019)Cao, Tang, Fang, Shen, Lu, and Tai]{cda}
Jinkun Cao, Hongyang Tang, Hao-Shu Fang, Xiaoyong Shen, Cewu Lu, and Yu-Wing
  Tai.
\newblock Cross-domain adaptation for animal pose estimation.
\newblock In \emph{Proceedings of the IEEE/CVF international conference on
  computer vision}, pages 9498--9507, 2019.

\bibitem[Del~Pero et~al.(2016)Del~Pero, Ricco, Sukthankar, and Ferrari]{tigdog}
L. Del~Pero, S. Ricco, R. Sukthankar, and V. Ferrari.
\newblock Behavior discovery and alignment of articulated object classes from
  unstructured video.
\newblock \emph{International Journal of Computer Vision (IJCV)}, 2016.

\bibitem[DESCRiPtoR()]{descriptorposes}
Data DESCRiPtoR.
\newblock the poses for equine research dataset (pferd).

\bibitem[Ebadi et~al.(2022)Ebadi, Dhakad, Vishwakarma, Wang, Jhang, Chociej,
  Crespi, Thaman, and Ganguly]{peoplesanspeople}
Salehe~Erfanian Ebadi, Saurav Dhakad, Sanjay Vishwakarma, Chunpu Wang,
  You-Cyuan Jhang, Maciek Chociej, Adam Crespi, Alex Thaman, and Sujoy Ganguly.
\newblock Psp-hdri+: A synthetic dataset generator for pre-training of
  human-centric computer vision models.
\newblock In \emph{First Workshop on Pre-training: Perspectives, Pitfalls, and
  Paths Forward at ICML 2022}, 2022.

\bibitem[GiM Studio, \url{https://gim.studio/animalia/}()]{gimstudio}
GiM Studio, \url{https://gim.studio/animalia/}.

\bibitem[Graving et~al.(2019)Graving, Chae, Naik, Li, Koger, Costelloe, and
  Couzin]{dpk-zebrads}
Jacob~M Graving, Daniel Chae, Hemal Naik, Liang Li, Benjamin Koger, Blair~R
  Costelloe, and Iain~D Couzin.
\newblock Deepposekit, a software toolkit for fast and robust animal pose
  estimation using deep learning.
\newblock \emph{eLife}, 8:\penalty0 e47994, 2019.

\bibitem[Higami et~al.(2024)Higami, Oshima, Shiramatsu, Takahashi, Nobuhara,
  and Nishino]{higami2024ratbodyformer}
Ayaka Higami, Karin Oshima, Tomoyo~Isoguchi Shiramatsu, Hirokazu Takahashi,
  Shohei Nobuhara, and Ko Nishino.
\newblock Ratbodyformer: Rat body surface from keypoints.
\newblock \emph{arXiv preprint arXiv:2412.09599}, 2024.

\bibitem[Jiang and Ostadabbas(2023)]{spacnet}
Le Jiang and Sarah Ostadabbas.
\newblock Spac-net: Synthetic pose-aware animal controlnet for enhanced pose
  estimation, 2023.

\bibitem[Jiang et~al.(2022{\natexlab{a}})Jiang, Lee, Teotia, and
  Ostadabbas]{surveyKPs}
Le Jiang, Caleb Lee, Divyang Teotia, and Sarah Ostadabbas.
\newblock Animal pose estimation: A closer look at the state-of-the-art,
  existing gaps and opportunities.
\newblock \emph{Computer Vision and Image Understanding}, 222:\penalty0 103483,
  2022{\natexlab{a}}.

\bibitem[Jiang et~al.(2022{\natexlab{b}})Jiang, Liu, Bai, and
  Ostadabbas]{pasyn}
Le Jiang, Shuangjun Liu, Xiangyu Bai, and Sarah Ostadabbas.
\newblock Prior-aware synthetic data to the rescue: Animal pose estimation with
  very limited real data.
\newblock In \emph{The British Machine Vision Conference (BMVC)},
  2022{\natexlab{b}}.

\bibitem[Jocher et~al.(2022)Jocher, Chaurasia, Stoken, and et. al.]{yolov5}
Glenn Jocher, Ayush Chaurasia, Alex Stoken, and et. al.
\newblock {ultralytics/yolov5: v7.0 - YOLOv5 SOTA Realtime Instance
  Segmentation}, 2022.

\bibitem[Kulits et~al.()Kulits, Feng, Liu, Abrevaya, and Black]{kulitsre}
Peter Kulits, Haiwen Feng, Weiyang Liu, Victoria~Fernandez Abrevaya, and
  Michael~J Black.
\newblock Re-thinking inverse graphics with large language models.
\newblock \emph{Transactions on Machine Learning Research}.

\bibitem[Kulits et~al.(2025)Kulits, Black, and Zuffi]{kulits2025reconstructing}
Peter Kulits, Michael~J Black, and Silvia Zuffi.
\newblock Reconstructing animals and the wild.
\newblock In \emph{Proceedings of the Computer Vision and Pattern Recognition
  Conference}, pages 16565--16577, 2025.

\bibitem[Li and Lee(2021)]{uda}
Chen Li and Gim~Hee Lee.
\newblock From synthetic to real: Unsupervised domain adaptation for animal
  pose estimation.
\newblock In \emph{Proceedings of the IEEE/CVF conference on computer vision
  and pattern recognition}, pages 1482--1491, 2021.

\bibitem[Li and Lee(2023)]{li2023scarcenet}
Chen Li and Gim~Hee Lee.
\newblock Scarcenet: Animal pose estimation with scarce annotations.
\newblock In \emph{Proceedings of the IEEE/CVF Conference on Computer Vision
  and Pattern Recognition}, pages 17174--17183, 2023.

\bibitem[Lin et~al.(2014)Lin, Maire, Belongie, Hays, Perona, Ramanan,
  Doll{\'a}r, and Zitnick]{cocodataset}
Tsung-Yi Lin, Michael Maire, Serge Belongie, James Hays, Pietro Perona, Deva
  Ramanan, Piotr Doll{\'a}r, and C.~Lawrence Zitnick.
\newblock Microsoft coco: Common objects in context.
\newblock In \emph{Computer Vision -- ECCV 2014}, pages 740--755, Cham, 2014.
  Springer International Publishing.

\bibitem[Liu et~al.(2023)Liu, Zeng, Ren, Li, Zhang, Yang, Li, Yang, Su, Zhu,
  et~al.]{groundingDino}
Shilong Liu, Zhaoyang Zeng, Tianhe Ren, Feng Li, Hao Zhang, Jie Yang, Chunyuan
  Li, Jianwei Yang, Hang Su, Jun Zhu, et~al.
\newblock Grounding dino: Marrying dino with grounded pre-training for open-set
  object detection.
\newblock \emph{arXiv preprint arXiv:2303.05499}, 2023.

\bibitem[Loper et~al.(2015)Loper, Mahmood, Romero, Pons-Moll, and Black]{smpl}
Matthew Loper, Naureen Mahmood, Javier Romero, Gerard Pons-Moll, and Michael~J.
  Black.
\newblock {SMPL}: A skinned multi-person linear model.
\newblock \emph{ACM Trans. Graph.}, 34\penalty0 (6), 2015.

\bibitem[Lyu et~al.(2025)Lyu, Zhu, Gu, Lin, Cheng, Liu, Tang, and
  An]{lyu2025animer}
Jin Lyu, Tianyi Zhu, Yi Gu, Li Lin, Pujin Cheng, Yebin Liu, Xiaoying Tang, and
  Liang An.
\newblock Animer: Animal pose and shape estimation using family aware
  transformer.
\newblock In \emph{Proceedings of the Computer Vision and Pattern Recognition
  Conference}, pages 17486--17496, 2025.

\bibitem[Mahmood et~al.(2019)Mahmood, Ghorbani, Troje, Pons-Moll, and
  Black]{amass}
Naureen Mahmood, Nima Ghorbani, Nikolaus~F. Troje, Gerard Pons-Moll, and
  Michael~J. Black.
\newblock {AMASS}: Archive of motion capture as surface shapes.
\newblock In \emph{International Conference on Computer Vision}, pages
  5442--5451, 2019.

\bibitem[Marshall et~al.(2021)Marshall, Aldarondo, Dunn, Wang, Berman, and
  {\"O}lveczky]{marshall2021continuous}
Jesse~D Marshall, Diego~E Aldarondo, Timothy~W Dunn, William~L Wang, Gordon~J
  Berman, and Bence~P {\"O}lveczky.
\newblock Continuous whole-body 3d kinematic recordings across the rodent
  behavioral repertoire.
\newblock \emph{Neuron}, 109\penalty0 (3):\penalty0 420--437, 2021.

\bibitem[Mu et~al.(2020)Mu, Qiu, Hager, and Yuille]{lsa}
Jiteng Mu, Weichao Qiu, Gregory~D Hager, and Alan~L Yuille.
\newblock Learning from synthetic animals.
\newblock In \emph{Proceedings of the IEEE/CVF Conference on Computer Vision
  and Pattern Recognition}, pages 12386--12395, 2020.

\bibitem[Nagy et~al.(2023)Nagy, Naik, Kano, Carlson, Koblitz, Wikelski, and
  Couzin]{nagy2023smart}
M{\'a}t{\'e} Nagy, Hemal Naik, Fumihiro Kano, Nora~V Carlson, Jens~C Koblitz,
  Martin Wikelski, and Iain~D Couzin.
\newblock Smart-barn: Scalable multimodal arena for real-time tracking behavior
  of animals in large numbers.
\newblock \emph{Science Advances}, 9\penalty0 (35):\penalty0 eadf8068, 2023.

\bibitem[Niewiadomski et~al.(2024)Niewiadomski, Yiannakidis, Cuevas-Velasquez,
  Sanyal, Black, Zuffi, and Kulits]{niewiadomski2024generative}
Tomasz Niewiadomski, Anastasios Yiannakidis, Hanz Cuevas-Velasquez, Soubhik
  Sanyal, Michael~J Black, Silvia Zuffi, and Peter Kulits.
\newblock Generative zoo.
\newblock \emph{CoRR}, 2024.

\bibitem[Price et~al.(2023)Price, Khandelwal, Rubenstein, and Ahmad]{mpala}
Eric Price, Pranav~C. Khandelwal, Daniel~I. Rubenstein, and Aamir Ahmad.
\newblock A framework for fast, large-scale, semi-automatic inference of animal
  behavior from monocular videos.
\newblock \emph{bioRxiv}, 2023.

\bibitem[PROMax3D, \url{https://www.promax3d.com/animal}()]{3danimalmodels}
PROMax3D, \url{https://www.promax3d.com/animal}.

\bibitem[Smith et~al.(2022)Smith, Gilbert, Woodfine, Kraaijeveld, Chege,
  Kimiti, Low-Mackey, Mutinda, Ngene, Rubenstein, Wandera, and
  Riordan]{Smith2022}
Chelsea~V. Smith, Tania~C. Gilbert, Tim Woodfine, Alex Kraaijeveld, Geoffrey
  Chege, David Kimiti, Belinda Low-Mackey, Mathew Mutinda, Shadrack Ngene, Dan
  Rubenstein, Anthony Wandera, and Philip Riordan.
\newblock Population and habitat connectivity of grevy’s zebra equus grevyi,
  a threatened large herbivore in degraded rangelands.
\newblock \emph{Biological Conservation}, 274:\penalty0 109711, 2022.

\bibitem[Trondrud et~al.(2022)Trondrud, Ugland, Ropstad, Loe, Albon, Stien,
  Evans, Thorsby, Veiberg, Irvine, and Pigeon]{Trondrud2022}
L.~Monica Trondrud, Cassandra Ugland, Erik Ropstad, Leif~Egil Loe, Steve Albon,
  Audun Stien, Alina~L. Evans, Per~Medbøe Thorsby, Vebjørn Veiberg, R.~Justin
  Irvine, and Gabriel Pigeon.
\newblock Stress responses to repeated captures in a wild ungulate.
\newblock \emph{Scientific Reports}, 12\penalty0 (1), 2022.

\bibitem[Xu et~al.(2024)Xu, Zhang, Zhang, and Tao]{vitposeplus}
Yufei Xu, Jing Zhang, Qiming Zhang, and Dacheng Tao.
\newblock Vitpose++: Vision transformer for generic body pose estimation.
\newblock \emph{IEEE Transactions on Pattern Analysis and Machine
  Intelligence}, 46\penalty0 (2):\penalty0 1212--1230, 2024.

\bibitem[Yang et~al.(2022)Yang, Yang, Xu, Zhang, Lan, and Tao]{apt-36k}
Yuxiang Yang, Junjie Yang, Yufei Xu, Jing Zhang, Long Lan, and Dacheng Tao.
\newblock Apt-36k: A large-scale benchmark for animal pose estimation and
  tracking.
\newblock In \emph{Advances in Neural Information Processing Systems}, pages
  17301--17313. Curran Associates, Inc., 2022.

\bibitem[Yu et~al.(2021)Yu, Xu, Zhang, Zhao, Guan, and Tao]{ap-10k}
Hang Yu, Yufei Xu, Jing Zhang, Wei Zhao, Ziyu Guan, and Dacheng Tao.
\newblock {AP}-10k: A benchmark for animal pose estimation in the wild.
\newblock In \emph{Thirty-fifth Conference on Neural Information Processing
  Systems Datasets and Benchmarks Track (Round 2)}, 2021.

\bibitem[Zhang et~al.(2023)Zhang, Wang, Chen, Xu, Zhang, and Tao]{clamp}
Xu Zhang, Wen Wang, Zhe Chen, Yufei Xu, Jing Zhang, and Dacheng Tao.
\newblock Clamp: Prompt-based contrastive learning for connecting language and
  animal pose.
\newblock In \emph{Proceedings of the IEEE/CVF Conference on Computer Vision
  and Pattern Recognition}, pages 23272--23281, 2023.

\bibitem[Zuffi et~al.(2017)Zuffi, Kanazawa, Jacobs, and Black]{smal}
Silvia Zuffi, Angjoo Kanazawa, David Jacobs, and Michael~J. Black.
\newblock {3D} menagerie: Modeling the {3D} shape and pose of animals.
\newblock In \emph{IEEE Conf. on Computer Vision and Pattern Recognition
  (CVPR)}, 2017.

\bibitem[Zuffi et~al.(2018)Zuffi, Kanazawa, and Black]{smalr}
Silvia Zuffi, Angjoo Kanazawa, and Michael~J. Black.
\newblock Lions and tigers and bears: Capturing non-rigid, {3D}, articulated
  shape from images.
\newblock In \emph{IEEE Conference on Computer Vision and Pattern Recognition
  (CVPR)}. IEEE Computer Society, 2018.

\bibitem[Zuffi et~al.(2019)Zuffi, Kanazawa, Berger-Wolf, and
  Black]{Zuffi2019ThreeDSL}
Silvia Zuffi, Angjoo Kanazawa, T. Berger-Wolf, and Michael~J. Black.
\newblock Three-d safari: Learning to estimate zebra pose, shape, and texture
  from images “in the wild”.
\newblock \emph{2019 IEEE/CVF International Conference on Computer Vision
  (ICCV)}, pages 5358--5367, 2019.

\end{thebibliography}
		}
		
		\clearpage
\setcounter{page}{1}
\maketitlesupplementary

\section*{Overview}
\begin{figure}[H]
\begin{minipage}{1.0\textwidth}
  \strut\newline
    \centering
 
    \subcaptionbox{Trained with $\text{CZ}_{1920}$}{\includegraphics[width=0.49\textwidth]{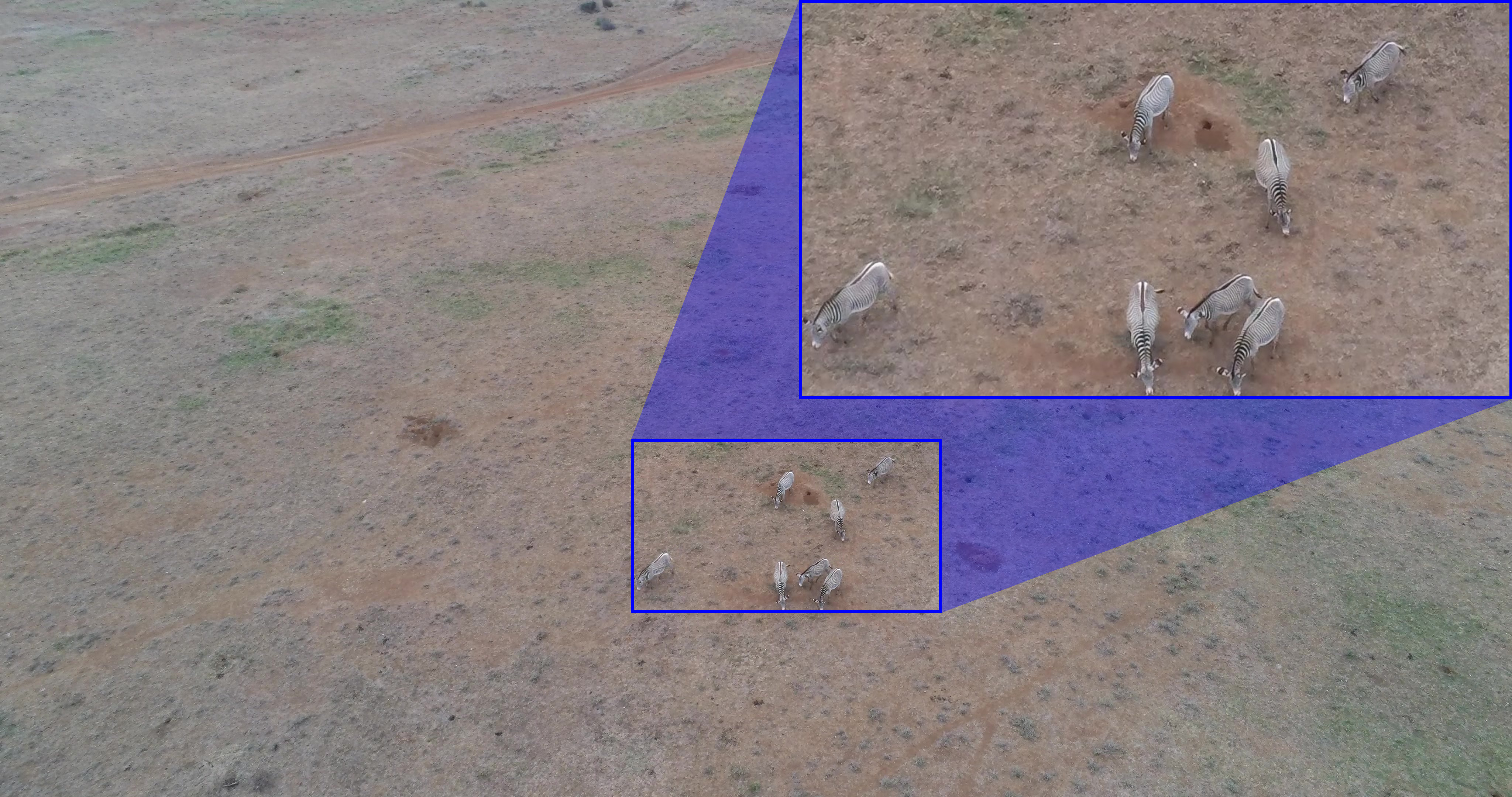}}\hfill
    \subcaptionbox{Trained with SC}{\includegraphics[width=0.49\textwidth]{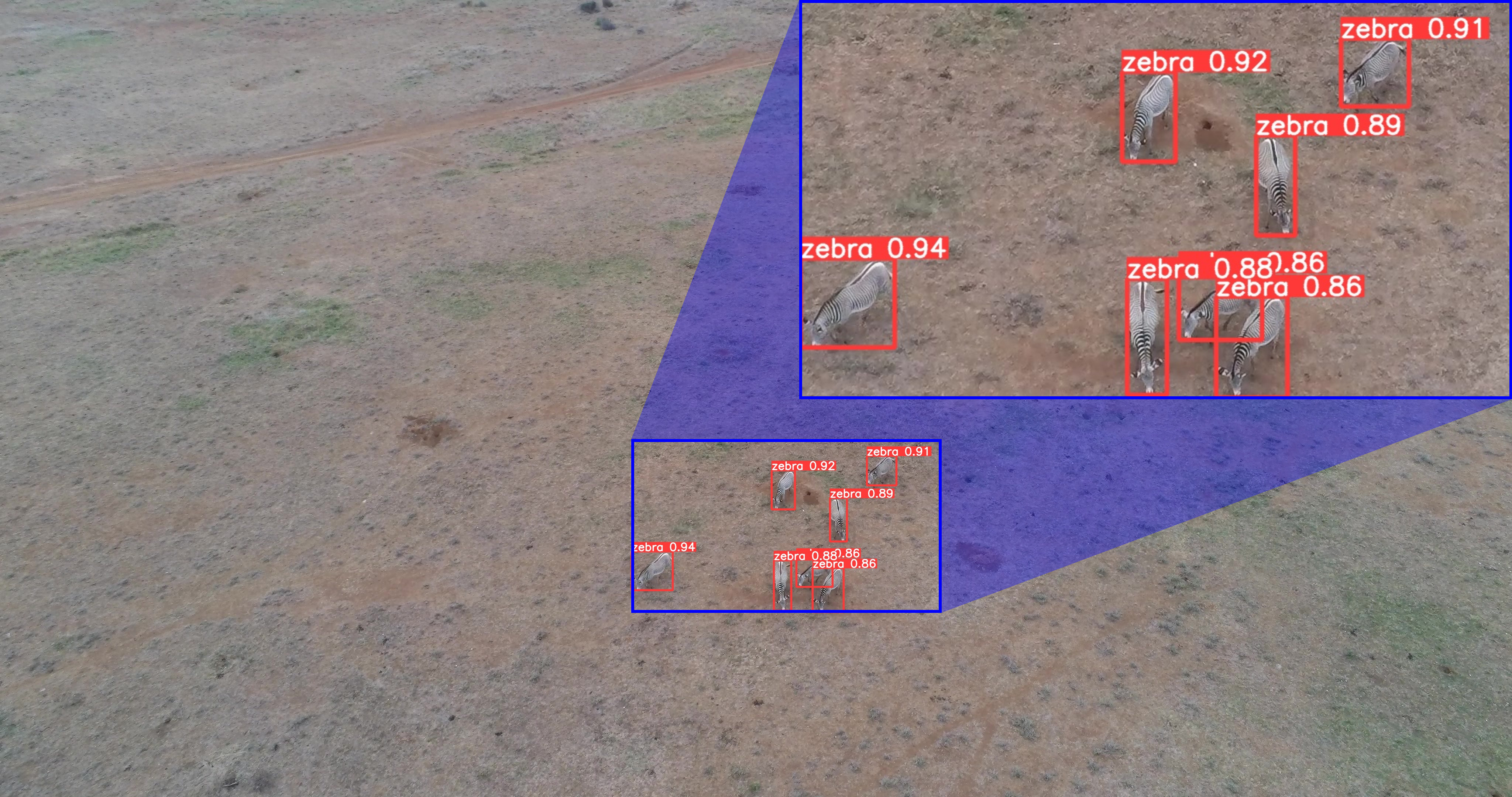}}\hfill     
    \subcaptionbox{Trained with $\text{SC}_\text{5K}$}{\includegraphics[width=0.49\textwidth]{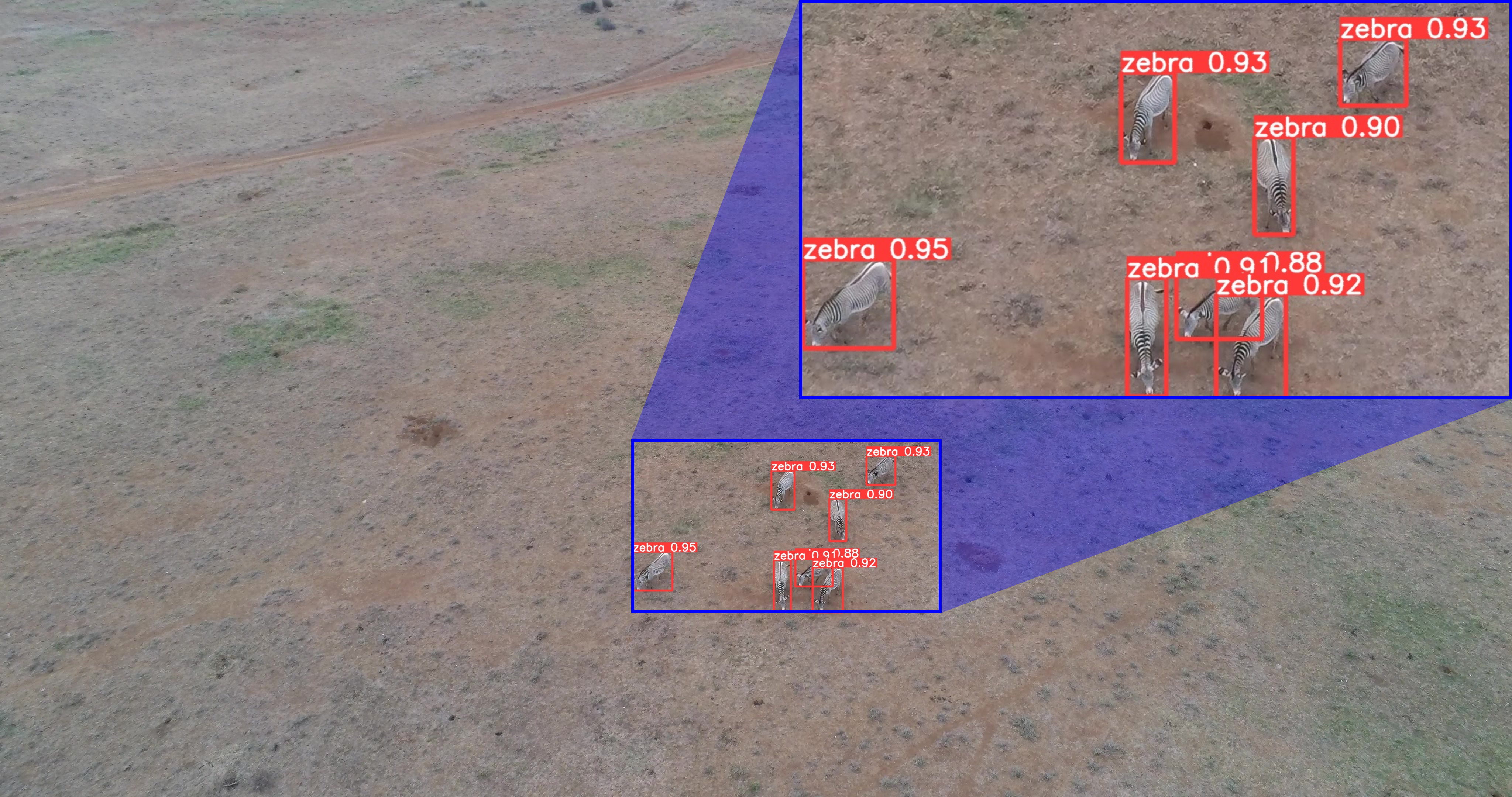}}\hfill
    \subcaptionbox{Trained with $\text{SC}_\text{5K}$+CZ$_{1920}$+RP}{\includegraphics[width=0.49\textwidth]{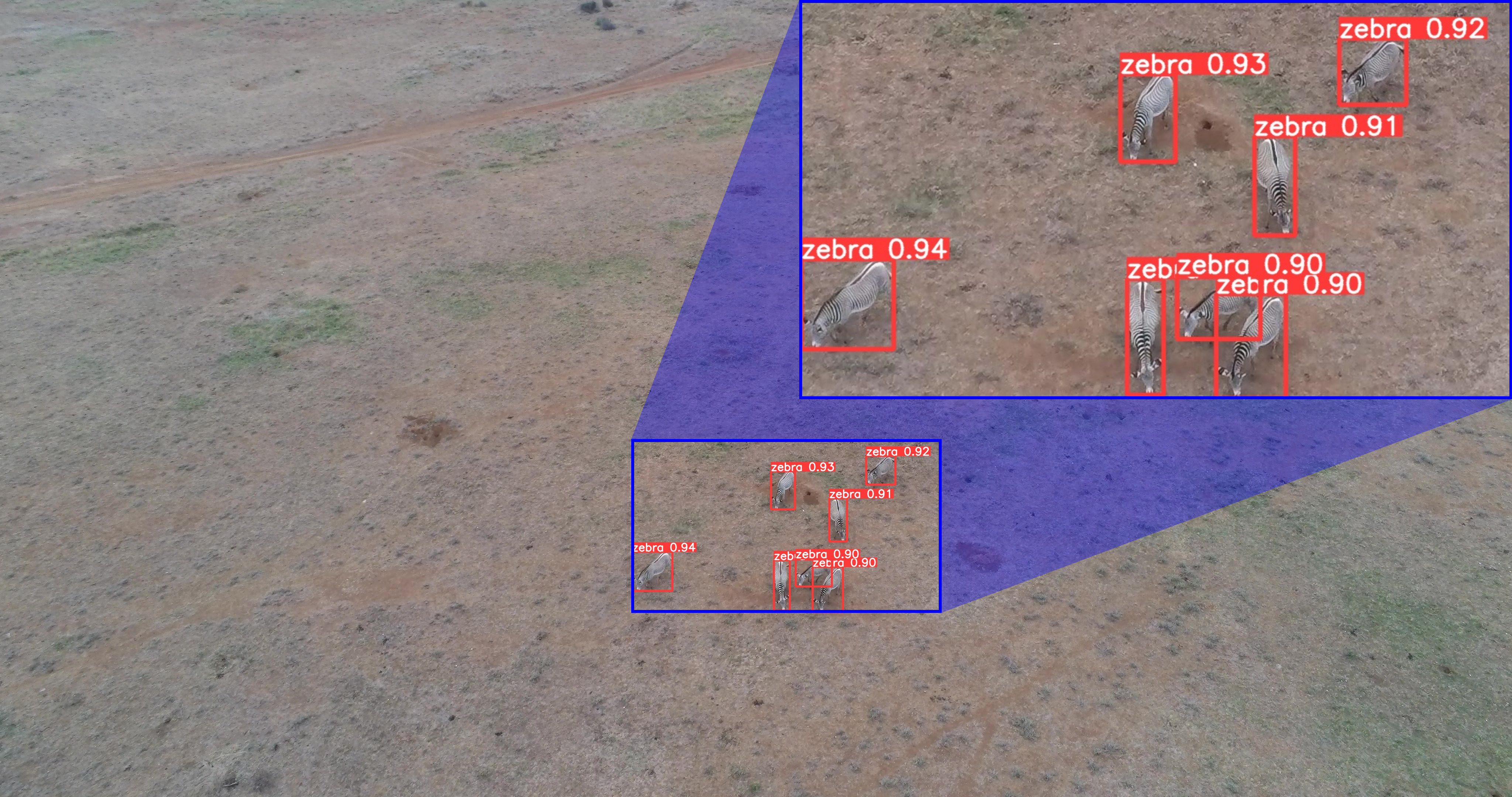}}\hfill
    
    \caption{YOLOv5s results on images taken from~\cite{mpala} with images scaled to 1920px.}
    \label{fig:qualitative-yolo-supp}
\end{minipage}
    \end{figure}

\begin{figure*}[]
    \centering
    \subcaptionbox{SC}{\includegraphics[width=0.49\textwidth]{fig/drone_scr/s/vis_0_crop.jpg}}\hfill
    \subcaptionbox{$\text{A10}_\text{OZ}$}{\includegraphics[width=0.49\textwidth]{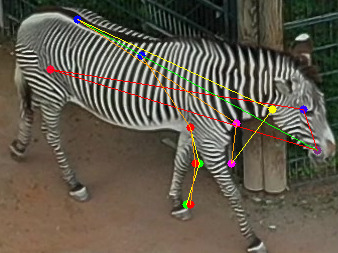}}\hfill
    \subcaptionbox{SC+$\text{A10}_\text{99}$}{\includegraphics[width=0.49\textwidth]{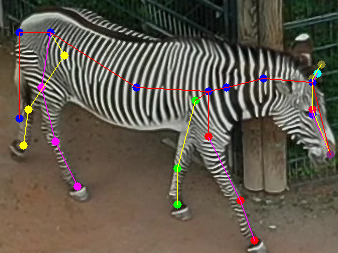}}\hfill
    \subcaptionbox{SC+$\text{A10}$}{\includegraphics[width=0.49\textwidth]{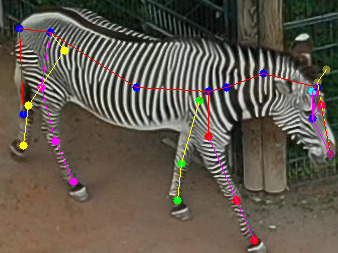}}\hfill
    \caption{ViTPose+ trained on the specified dataset using a randomly initialized backbone and run on one of the images from the R123 dataset, manually cropped around the zebra \textit{after} inference.}
    \label{fig:vit_drone_scr-supp}
\end{figure*}

\begin{figure*}[]
    \centering
    \subcaptionbox{SC}{\includegraphics[width=0.49\textwidth]{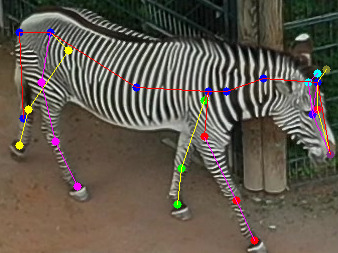}}\hfill
    \subcaptionbox{$\text{A10}_\text{OZ}$}{\includegraphics[width=0.49\textwidth]{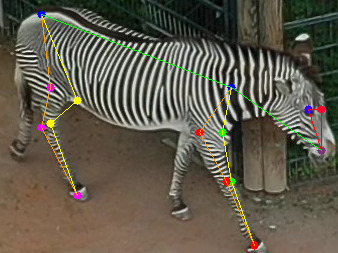}}\hfill
    \subcaptionbox{SC+$\text{A10}_\text{99}$}{\includegraphics[width=0.49\textwidth]{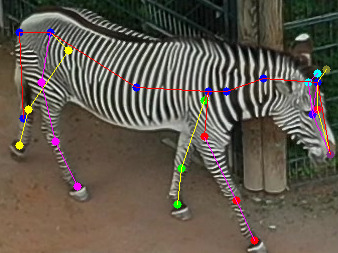}}\hfill
    \subcaptionbox{SC+$\text{A10}$}{\includegraphics[width=0.49\textwidth]{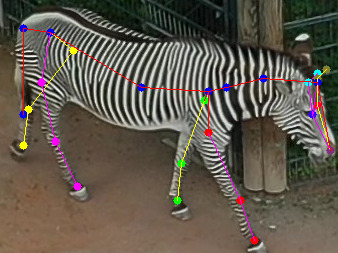}}\hfill
    \caption{ViTPose+ trained on the specified dataset using a MAE pre-trained backbone and run on one of the images from the R123 dataset, manually cropped around the zebra \textit{after} inference.}
    \label{fig:vit_drone_pre-supp}
\end{figure*}

\begin{figure*}[]
    \centering
    \subcaptionbox{SC}{\includegraphics[width=0.49\textwidth]{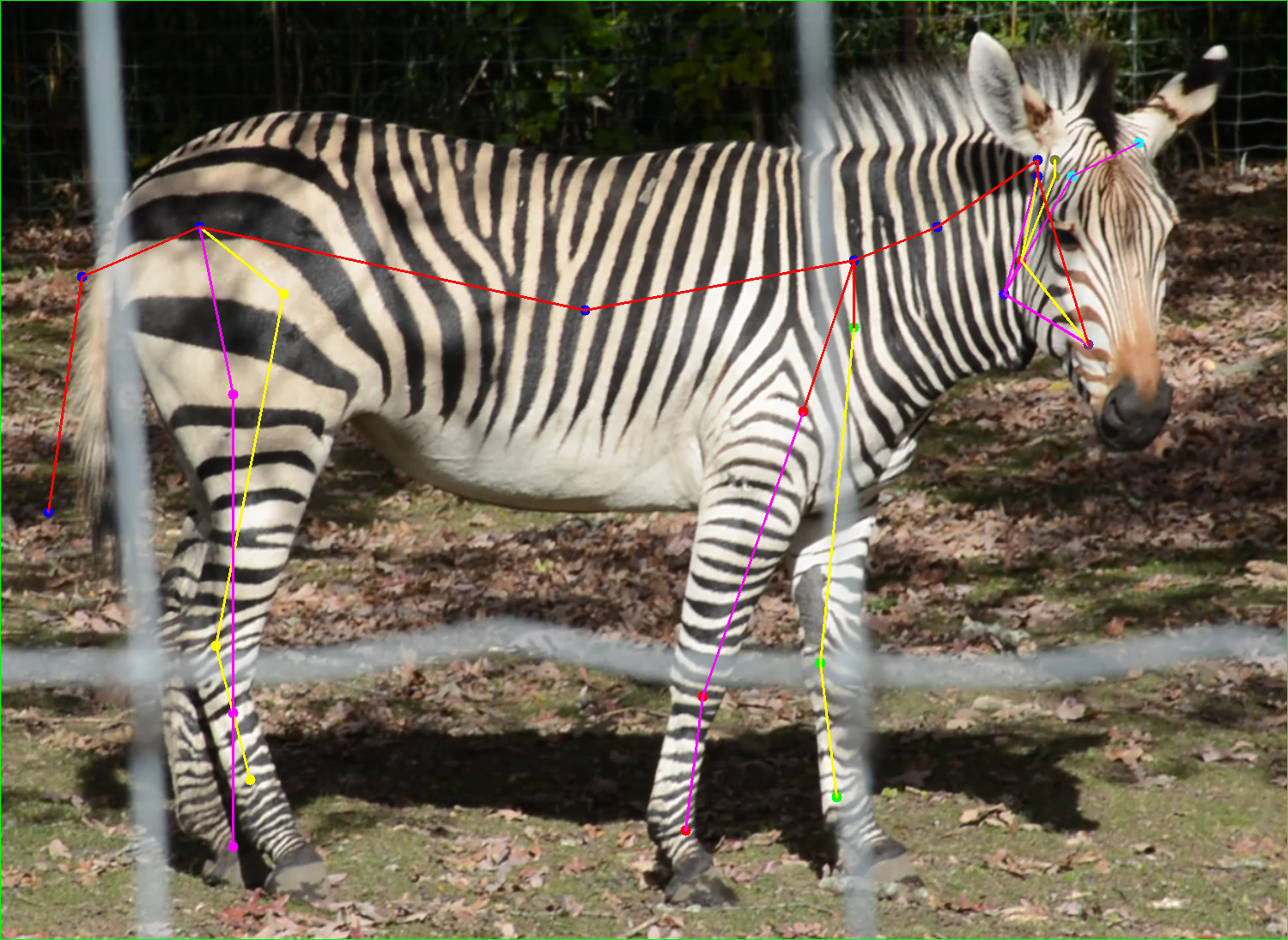}}\hfill
    \subcaptionbox{$\text{A10}_\text{OZ}$}{\includegraphics[width=0.49\textwidth]{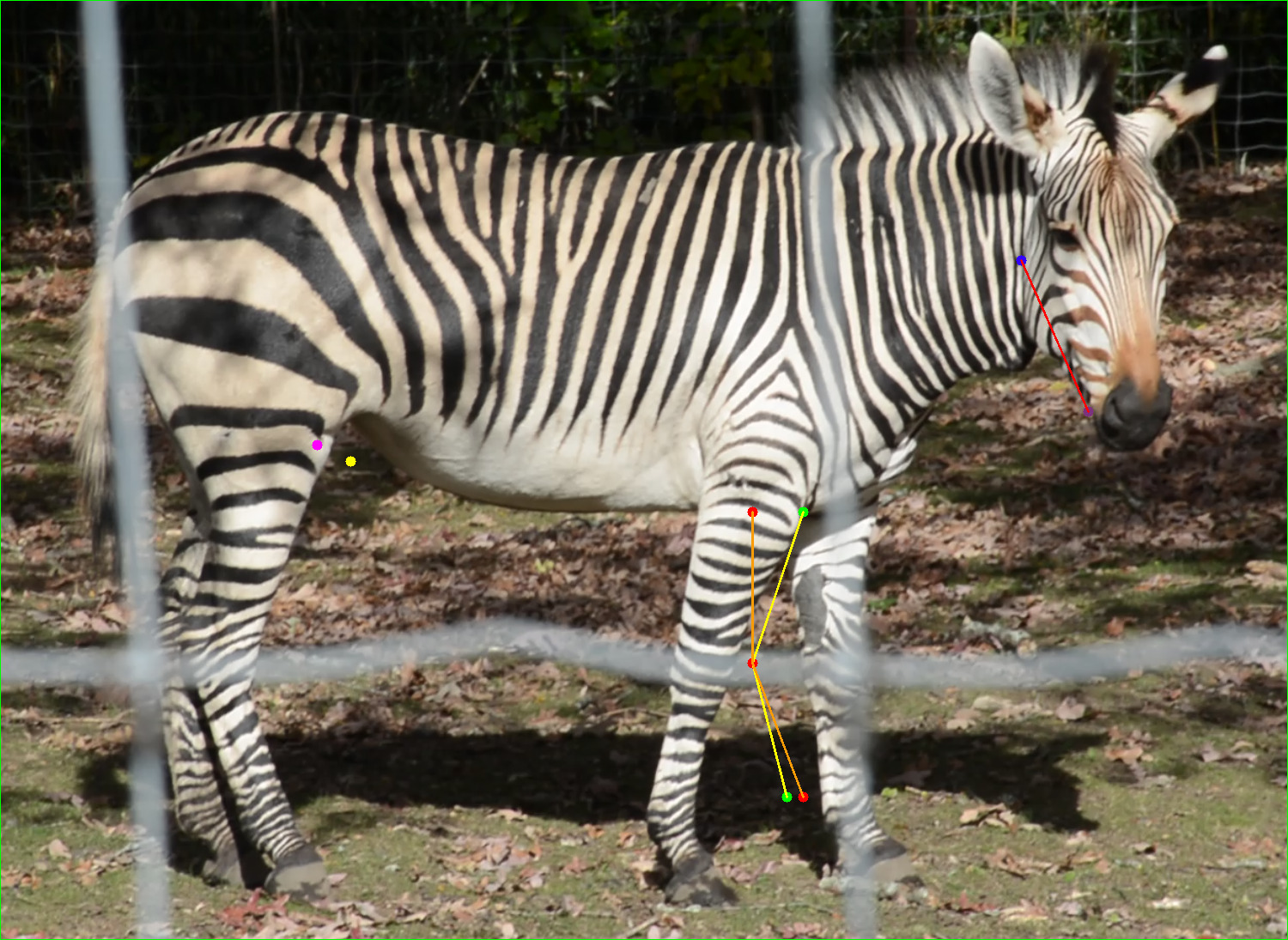}}\hfill
    
    \subcaptionbox{SC+$\text{A10}_\text{99}$}{\includegraphics[width=0.49\textwidth]{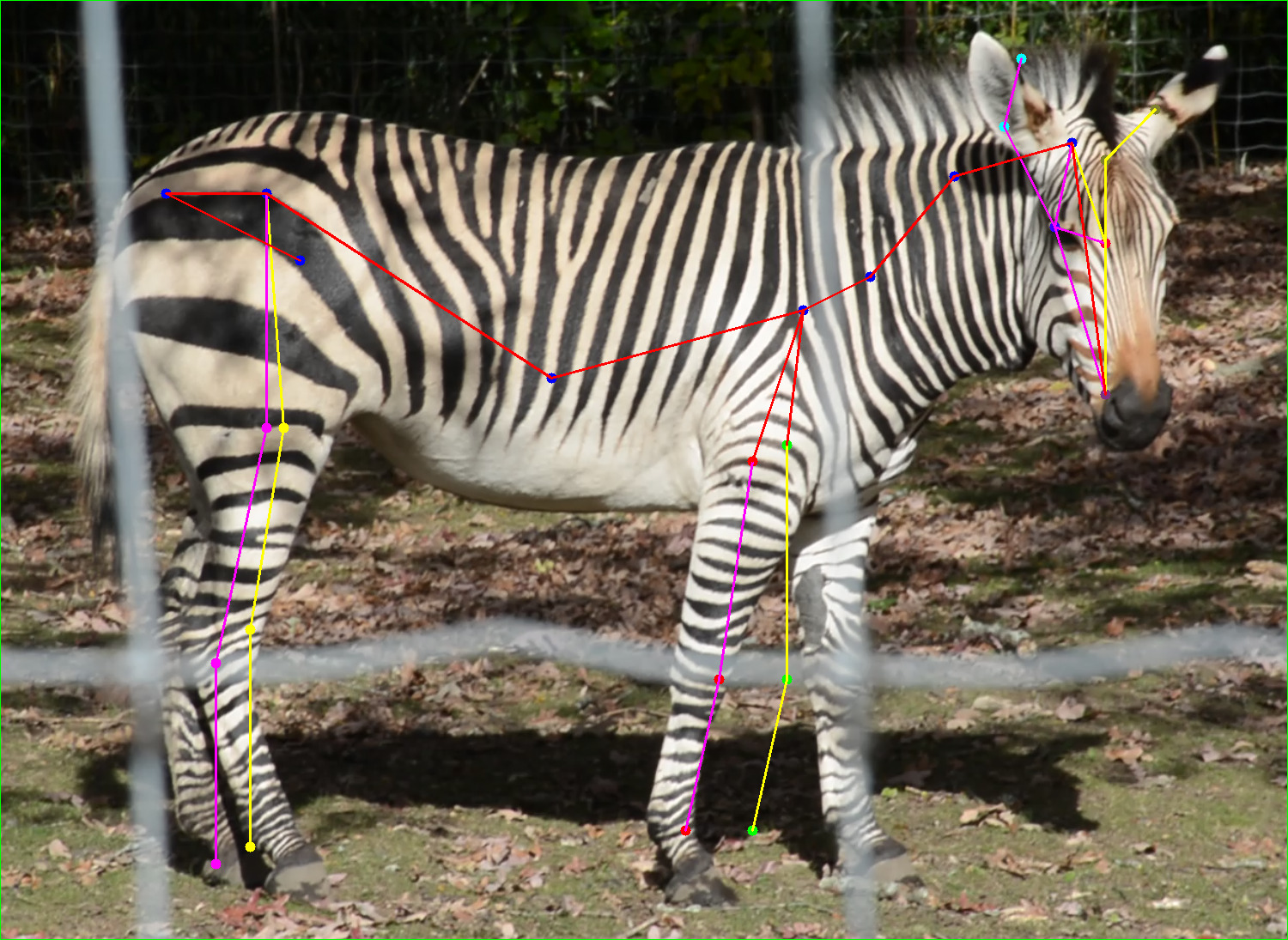}}\hfill
    \subcaptionbox{SC+$\text{A10}$}{\includegraphics[width=0.49\textwidth]{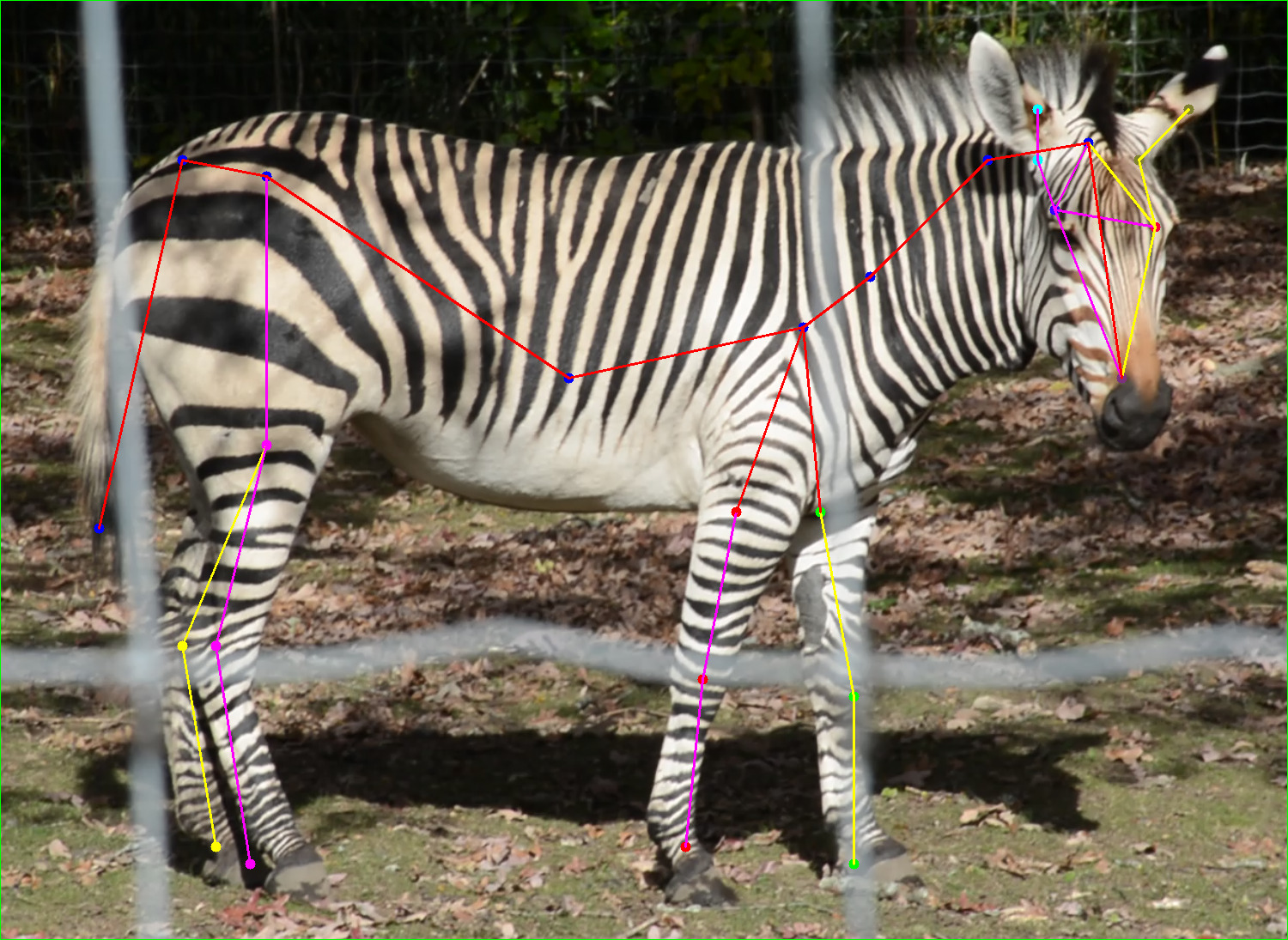}}\hfill
    \caption{ViTPose+ trained on the specified dataset using a randomly-initialized backbone and run on one of the images from the Zebra-zoo dataset, manually cropped around the zebra \textit{after} inference.}
    \label{fig:vit_zoo_no-pre-supp}
\end{figure*}

\begin{figure*}[]
    \centering
    \subcaptionbox{SC}{\includegraphics[width=0.49\textwidth]{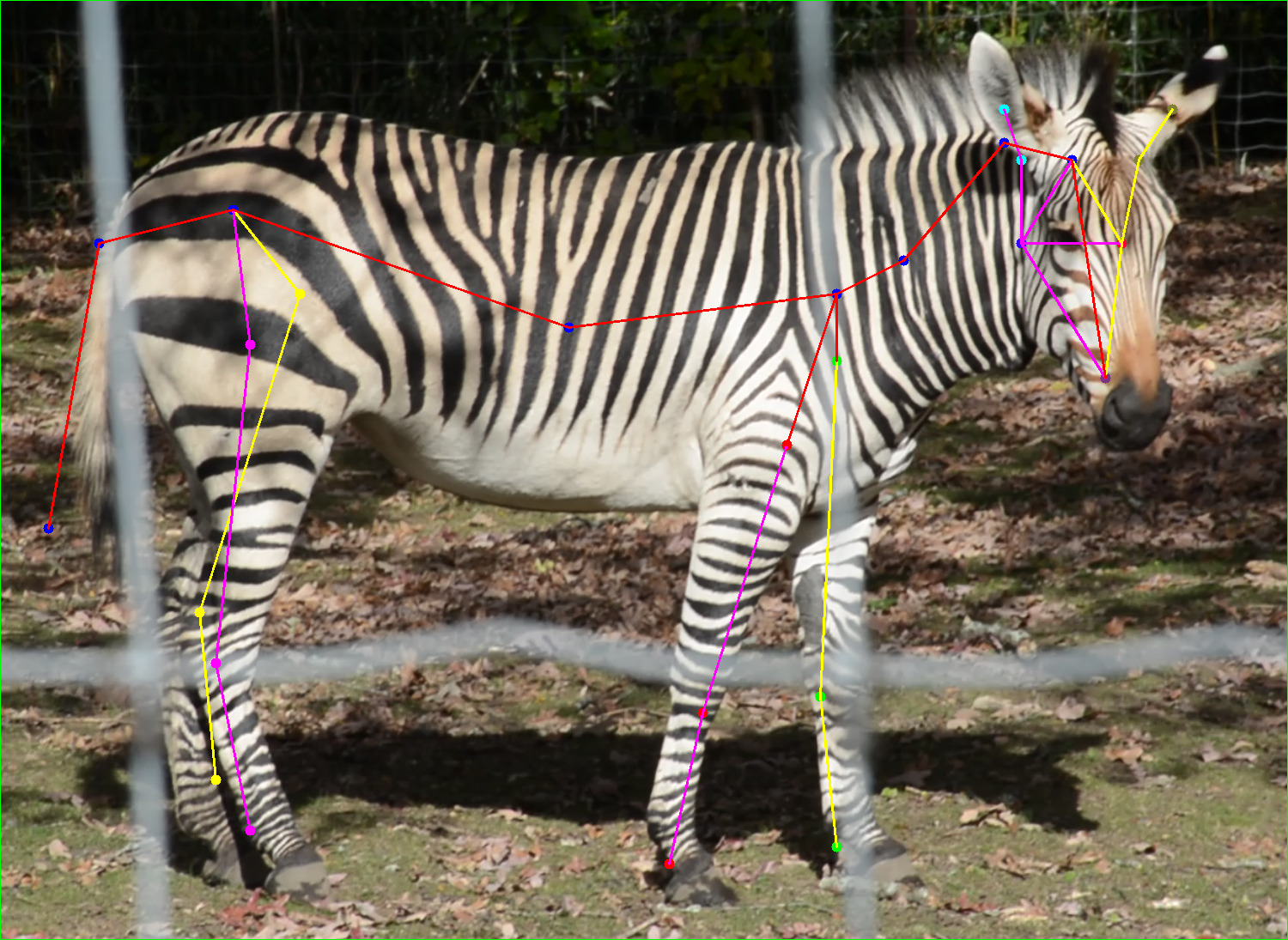}}\hfill
    \subcaptionbox{$\text{A10}_\text{OZ}$}{\includegraphics[width=0.49\textwidth]{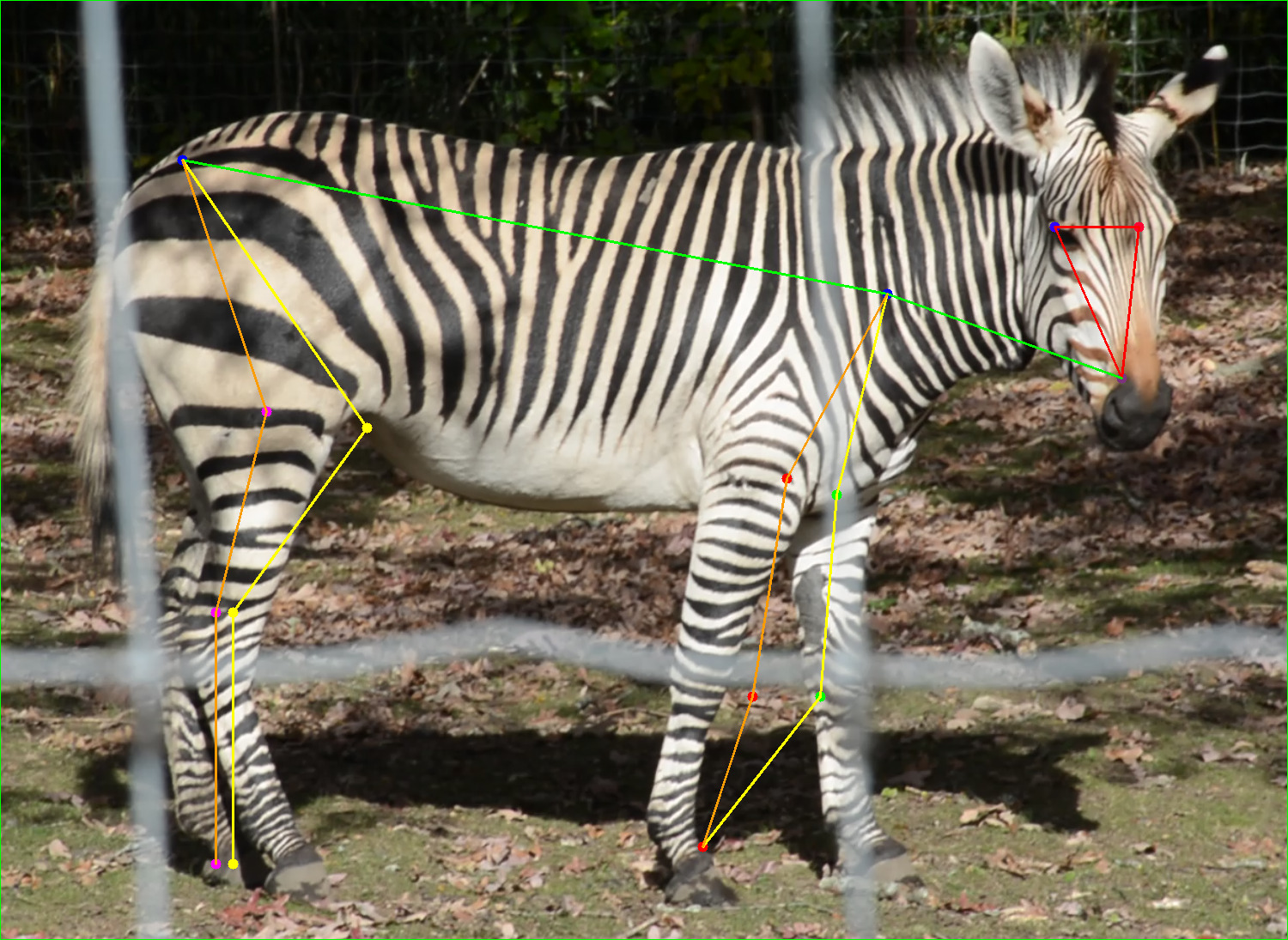}}\hfill
    
    \subcaptionbox{SC+$\text{A10}_\text{99}$}{\includegraphics[width=0.49\textwidth]{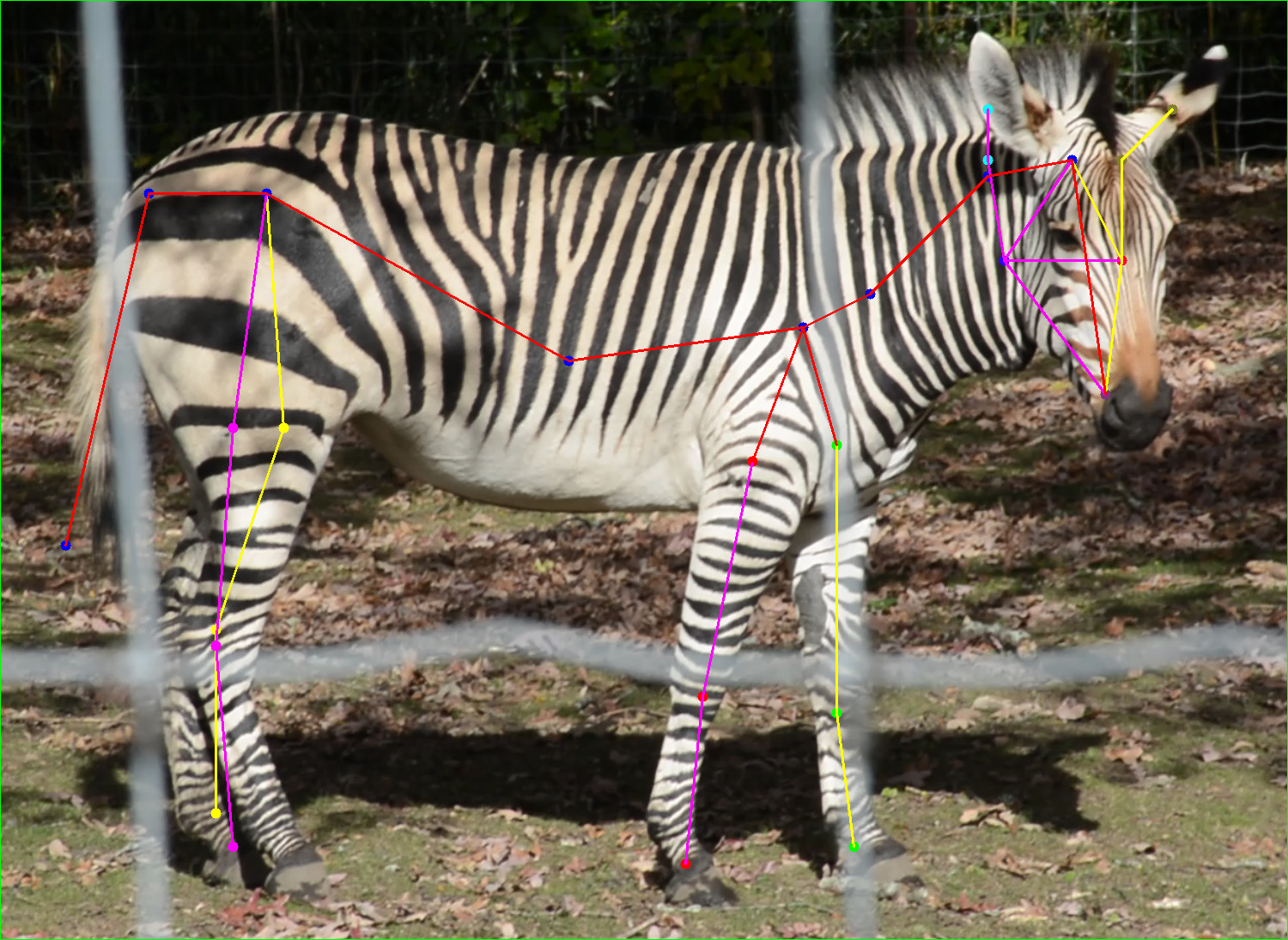}}\hfill
    \subcaptionbox{SC+$\text{A10}$}{\includegraphics[width=0.49\textwidth]{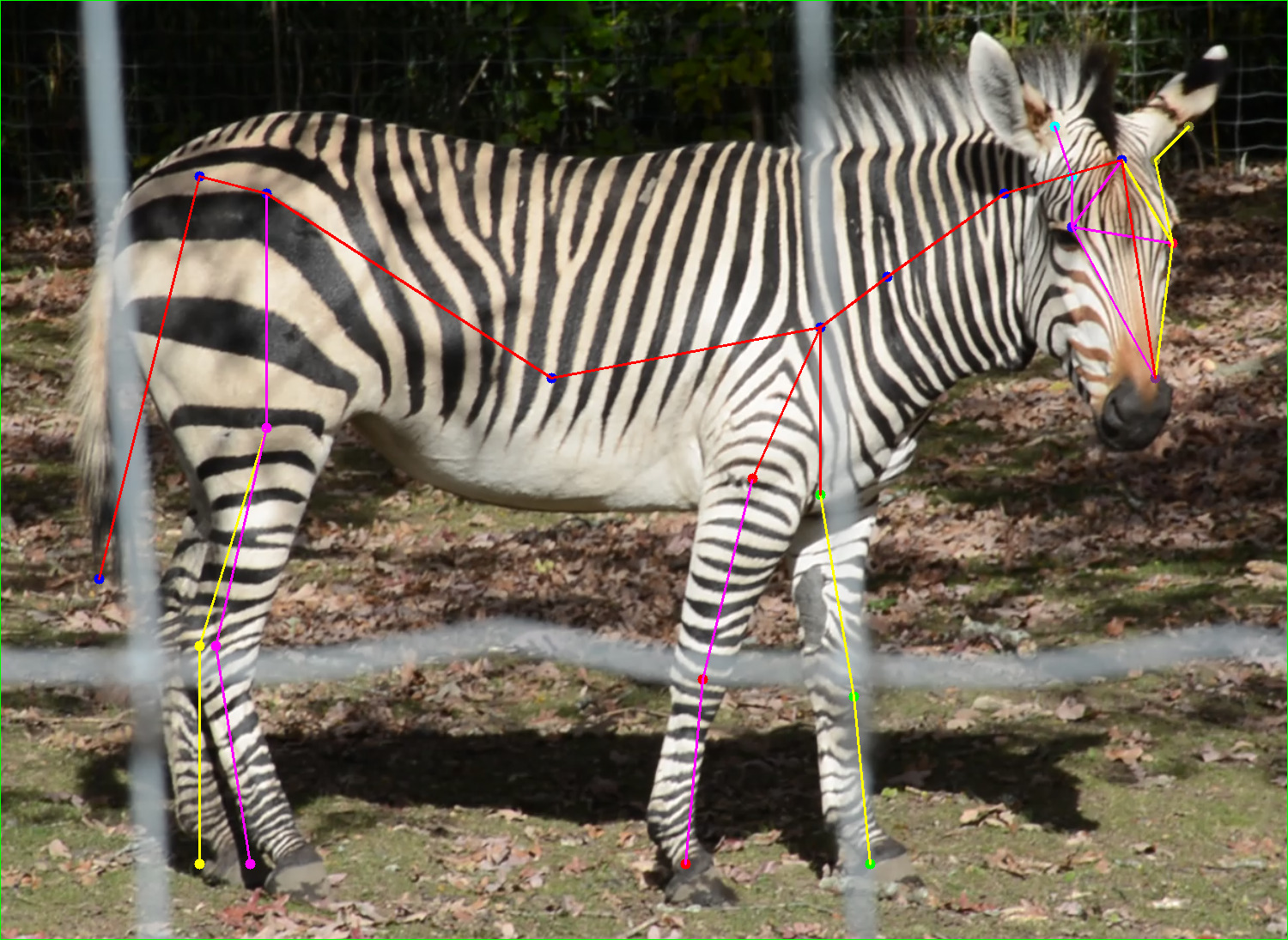}}\hfill
    \caption{ViTPose+ trained on the specified dataset using a MAE pre-trained backbone and run on one of the images from the Zebra-zoo dataset, manually cropped around the zebra \textit{after} inference.}
    \label{fig:vit_zoo_pre-supp}
\end{figure*}

\begin{figure*}[]
    \centering
    \subcaptionbox{SC}{\includegraphics[width=0.49\textwidth]{fig/horses_scratch/syn/vis_0.jpg}}\hfill
    \subcaptionbox{$\text{A10}_\text{OZ}$}{\includegraphics[width=0.49\textwidth]{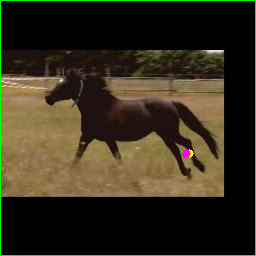}}\hfill
    \subcaptionbox{SC+$\text{TDH}_\text{99}$}{\includegraphics[width=0.49\textwidth]{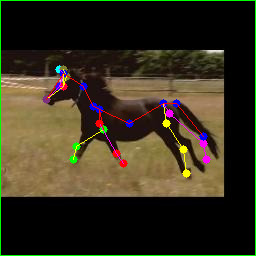}}\hfill
    \subcaptionbox{SC+$\text{TDH}$}{\includegraphics[width=0.49\textwidth]{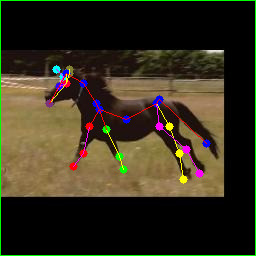}}\hfill
    \caption{ViTPose+ trained on the specified dataset using a randomly-initialized backbone and run on one of the images from the TDH dataset shown as \textit{processed} per dataset specifics.}
    \label{fig:vit_horse_scra-supp}
\end{figure*}

\begin{figure*}[]
    \centering
    \subcaptionbox{SC}{\includegraphics[width=0.49\textwidth]{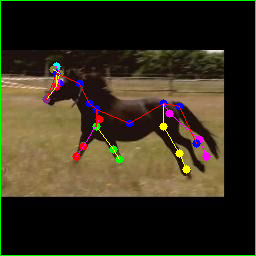}}\hfill
    \subcaptionbox{$\text{A10}_\text{OZ}$}{\includegraphics[width=0.49\textwidth]{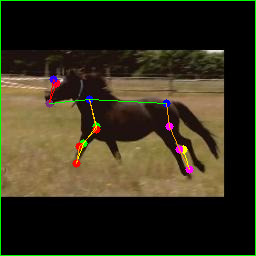}}\hfill
    \subcaptionbox{SC+$\text{TDH}_\text{99}$}{\includegraphics[width=0.49\textwidth]{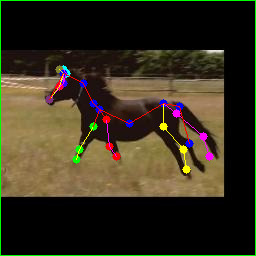}}\hfill
    \subcaptionbox{SC+$\text{TDH}$}{\includegraphics[width=0.49\textwidth]{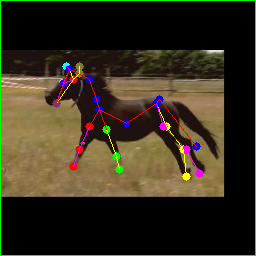}}\hfill
    \caption{ViTPose+ trained on the specified dataset using a MAE pre-trained backbone and run on one of the images from the TDH dataset shown as \textit{processed} per dataset specifics.}
    \label{fig:vit_horse_pre-supp}
\end{figure*}

\begin{table*}[!ht]
\centering
\resizebox{\textwidth}{!}{
\begin{tabular}{l|cc|c?ccc|cc|c|c|cc|c?cc}
   \toprule[1.5pt]
 & \multicolumn{3}{c?}{Synthetic} & \multicolumn{10}{c?}{ \begin{tabular}{c}
      Real, Common
 \end{tabular}} & \multicolumn{2}{c}{\begin{tabular}{c}
      Real, Aerial
       \end{tabular}}\\ \cline{2-16}
 & SC & $\text{SC}_{\text{5K}}$ & SpacNet & A10 & $\text{A10}_\text{OZ}$ & $\text{A10}_{99}$ & A36 & $\text{A36}_\text{OZ}$ & Zebra-300 & Zebra-Zoo & TDH & $\text{TDH}_{99}$ & $\text{CZ}_{[1920,640]}$ & R123 & RP \\
  & ~\cite{bonetto-syn-zebras} & \textbf{New} & ~\cite{spacnet} & \multicolumn{3}{c|}{~\cite{ap-10k}} & \multicolumn{2}{c|}{~\cite{apt-36k}} & ~\cite{pasyn} & ~\cite{pasyn} & \multicolumn{2}{c|}{~\cite{tigdog}} & ~\cite{cocodataset} &  & \\ \hline
Train & 14401 & 23184 & 2640 & 7023 & 140 & 80 & 28457 & 960 & --- & --- & 8380 & 80 & 1916 & --- & 720 \\
Valid & 3599 & 5798 & 360 & 995 & 20 & 19 & 7026 & 240 & 300 & 100 & 1772 & 19 & 85 & 104K  & 185 \\
Test & --- & --- & --- & 1997 & 40& ---& --- & --- & --- & --- & --- & --- & --- & ---  & --- \\ \hline
 Animal & Zebras & Zebras & Zebras & Various & Zebras & Zebras & Various & Zebras & Zebras & Zebras & Horses & Horses & Zebras & Zebras & Zebras \\ 
 \bottomrule[1.5pt]
\end{tabular}
}
\caption{Datasets used in this work. We indicate the number of images in each train/validation set and the animal(s) included.}
\label{tab:datasets}
\end{table*}
This is the supplementary material for ZebraPose. In \Cref{tab:datasets}, we provide a summary of the training and validation datasets. In there, we include the number of images for the training and validation sets, along with the types of animals contained in each dataset used in this work. Additionally, we present several qualitative results to further illustrate the performance of our models across different datasets and training configurations in~\Cref{fig:qualitative-yolo-supp} for YOLO, and~\Cref{fig:vit_drone_scr-supp,fig:vit_drone_pre-supp,fig:vit_zoo_no-pre-supp,fig:vit_zoo_pre-supp,fig:vit_horse_scra-supp,fig:vit_horse_pre-supp} for ViTPose. All the data (synthetic and real), network weights, results, and code will be open-source.


	\end{document}